\documentclass[sigconf]{acmart}

\usepackage{amsmath,amssymb,graphicx,hyperref}
\usepackage{booktabs}
\usepackage{adjustbox}
\usepackage{graphicx}
\usepackage{subcaption}   

\usepackage{balance}
\usepackage[htt]{hyphenat}   
\usepackage{microtype}       
\usepackage{flushend}  
\allowdisplaybreaks  
\AtBeginDocument{%
  }

\begin{document}


\title{Towards Platonic Representation for Table Reasoning: A Foundation for Permutation-Invariant Retrieval}
%

  

\author{Willy Carlos Tchuitcheu}
\authornote{Corresponding author (willy.Carlos.tchuitcheu@vub.be).}
\affiliation{
  \institution{Department of Mathematics and Data Science, Vrije Universiteit Brussel (VUB)}
  \city{Brussels}
  \country{Belgium}
}

\author{Tan Lu}
\affiliation{
  \institution{Department of Mathematics and Data Science, Vrije Universiteit Brussel (VUB)}
  \city{Brussels}
  \country{Belgium}
}
\affiliation{
  \institution{Data Science Lab, Royal Library of Belgium (KBR)}
  \city{Brussels}
  \country{Belgium}
}

\author{Ann Dooms}
\affiliation{
  \institution{Department of Mathematics and Data Science, Vrije Universiteit Brussel (VUB)}
  \city{Brussels}
  \country{Belgium}
}

\setcopyright{none}
\settopmatter{printacmref=false}
\renewcommand\footnotetextcopyrightpermission[1]{}
\pagestyle{plain}
\renewcommand{\shortauthors}{WC. Tchuitcheu et al.}

\begin{abstract}
Historical approaches to Table Representation Learning (TRL) have largely adopted the sequential paradigms of Natural Language Processing (NLP). We argue that this linearization of tables discards their essential geometric and relational structure, creating representations that are brittle to layout permutations. This paper introduces the Platonic Representation Hypothesis (PRH) for tables, positing that a semantically robust latent space for table reasoning must be intrinsically Permutation-Invariant (PI). To ground this hypothesis, we first conduct a retrospective analysis of table-reasoning tasks, highlighting the pervasive "serialization bias" that compromises structural integrity. We then propose a formal framework to diagnose this bias, introducing two principled metrics based on Centered Kernel Alignment (CKA): (i) $\mathrm{PI}_{\text{derange}}$, which measures embedding drift under complete structural derangement, and (ii) $\rho_{\text{mono}}$, a Spearman-based metric that tracks the convergence of latent structures toward a canonical form as structural information is incrementally restored. Our empirical analysis quantifies an expected flaw in modern Large Language Models (LLMs): even minor layout permutations induce significant, disproportionate semantic shifts in their table embeddings. This exposes a fundamental vulnerability in Retrieval-Augmented Generation (RAG) systems, in which table retrieval becomes fragile to layout-dependent noise rather than to semantic content. In response, we present a novel, structure-aware TRL encoder architecture that explicitly enforces the cognitive principle of cell–header alignment. This model demonstrates superior geometric stability and moves towards the PI ideal. Our work provides both a foundational critique of linearized table encoders and the theoretical scaffolding for semantically stable, permutation-invariant retrieval, charting a new direction for table reasoning in information systems.
\end{abstract}


\keywords{Table Representation Learning, Permutation Invariance, Platonic Representation Hypothesis, Evaluation Metrics, Semantic Drift, Table Reasoning, Information Retrieval}


\maketitle

\begin{figure*}
    \centering
    \includegraphics[height=5.5cm, width=1\linewidth]{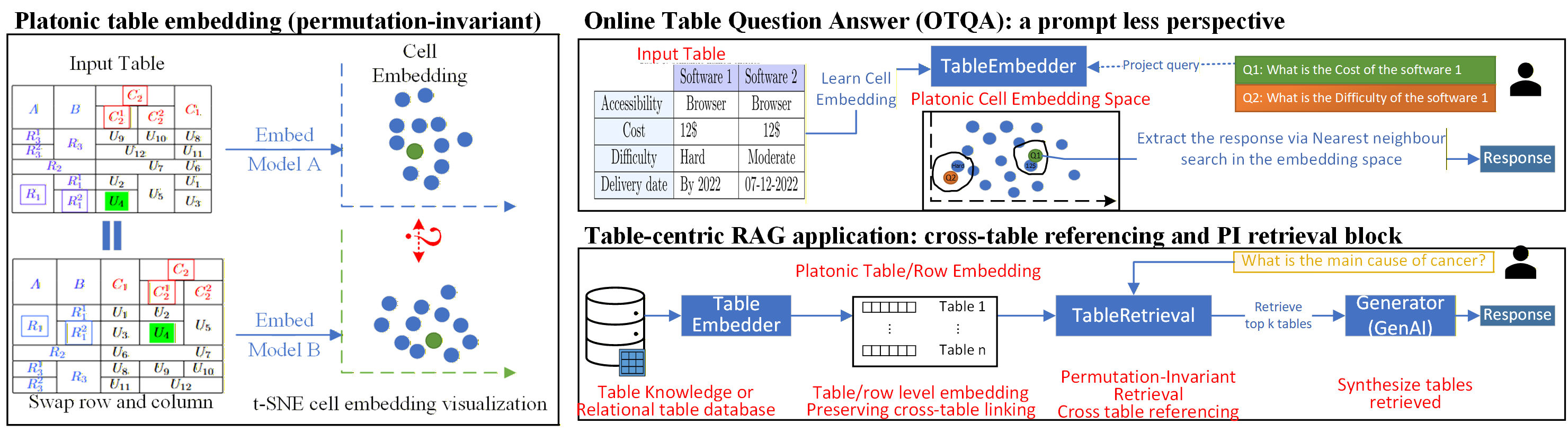}
\caption{Platonic view of permutation invariance in table embeddings and prospective long-term impact paradigm shift in  IR. The left part illustrates the theoretical concept of "Platonic table embedding," investigating whether an original input table and its structurally transposed counterpart (with rows and columns swapped) can be projected by embedding models into a consistent, semantically invariant latent space. The t-SNE visualizations and the question mark between them highlight the core research question regarding the alignment of these embedding spaces across different table representations. The right panel depicts prospective downstream applications that could be built upon such robust embeddings. The top section shows an "Online Table Question Answer" system that uses cell embeddings and nearest-neighbor search to directly answer user queries, while the bottom section outlines a "table-centric RAG application" that leverages table vectors for RAG tasks.}
    \label{fig:contributions}
\end{figure*}

\section{Introduction}
\label{sec:intro}


Information Retrieval (IR) with Retrieval-Augmented Generation (RAG) \cite{lewis2020retrieval} is undergoing a major shift: rather than generating answers only from retrieved unstructured text chunks, modern systems increasingly aim to produce concise answers by synthesizing evidence from heterogeneous and (semi-)structured sources such as tables \cite{chen2024tablerag}. However, the strong performance of Large Language Models (LLMs) on plain text has not translated into comparable reliability on (semi-)structured inputs. Recent table reasoning benchmarks and systems show that LLMs still struggle to consistently interpret and reason over tables, which remains a key bottleneck for interactive systems built on large-scale structured knowledge \cite{wu2025tablebench,su2025bright}.

Tables are not just “text in cells.” They encode relationships, hierarchies, and strict structure. A key property is row–column permutation invariance (PI): permuting rows or columns should not change the table’s meaning. Conventional dense retrieval, however, typically fails to preserve this property. Because retrieval is the foundation of any RAG system, the quality and structure of the vector representations directly affect what context is retrieved and, in turn, how well the system can reason. Recent evaluations, including BRIGHT \cite{su2025bright} and TableRAG \cite{chen2024tablerag}, reveal a major limitation: modern dense retrievers struggle with reasoning-heavy queries, where surface-level semantic similarity does not match true logical or relational relevance. A major cause is the common practice of serializing tables into linear text strings.

In this work, we first empirically quantify how such transformations, especially linearization, erase a table’s structural properties, with a focus on PI. We show that this issue persists even when using LLMs fine-tuned for tabular tasks, suggesting a deeper limitation of current approaches. This aligns with recent evidence that LLM performance on table understanding has started to plateau, as observed in TableBench \cite{wu2025tablebench}. Motivated by these findings, we argue for Platonic representations: structure-aware table embeddings that explicitly preserve key invariants. We present these representations as a necessary step toward a dual-pathway IR vision (Figure \ref{fig:contributions}). First, we introduce an Online Table Question Answering (OTQA) setting, where retrieval-style lookup questions can be answered directly in embedding space, enabling fast, zero-shot responses without expensive prompting. Second, we argue that invariant-aware representations are crucial for practical IR workloads in industry, including multi-step aggregation, relational filtering, and cross-table reasoning.

The idea of “Platonic” representations has recently been introduced in representation learning, not as a fixed architecture, but as a hypothesis. The Platonic Representation Hypothesis (PRH) \cite{pmlr-v235-huh24a} is assumed to reflect the underlying representation that stays invariant to specific sources of variation. A core idea in PRH is that different views of the same entity are only different observations—or “shadows”—of a single underlying essence.  We adopt this perspective for Table Representation Learning (TRL) and ask a fundamental question: Do current table embeddings capture a “Platonic” representation that is invariant to row and column permutations? Under this view, any row- or column-permuted version of a table is simply another observation of the same table (see the example in the left panel of Figure ~\ref{fig:contributions}).

The remainder of this paper is structured as follows: Section~\ref{sec:contribution} provides an overview of our empirical investigation into \textit{linearization bias} and a vision for a paradigm shift in IR. Section~\ref{sec:background} provides relevant background. In Section~\ref{sec:method}, we detail our proposed methodology for diagnosing the serialization bias, followed by our experimental framework and results in Section~\ref{sec:experiments}. A dedicated prospective analysis is presented in Section~\ref{sec:prospective}, where we explore the implications of our findings for the future of IR, before concluding the paper in Section~\ref{sec:conclusion}.

\section{Overview}
\label{sec:contribution}
This prospective paper argues that \textit{permutation invariance (PI)} is a pinpoint yet overlooked property for table-centric RAG: if a representation cannot remain stable under row/column permutations, it is structurally misaligned with table reasoning. We therefore assess the PI of cell-level contextual embeddings and use the resulting empirical observations to motivate a forward-looking research direction. Our contributions are threefold:

\begin{itemize}
    \item \textbf{A diagnostic framework via the Platonic lens:} We instantiate the \emph{Platonic Representation Hypothesis} (PRH) for tabular data and formalize the notion that row/column permutations are ``shadows'' of the same table. Concretely, we evaluate the stability of cell-level embeddings across different serialization manifolds under controlled permutations.

    \item \textbf{Structural invariance metrics ($PI_{\text{derange}}$ \& $\rho_{\text{mono}}$):} We introduce a CKA-based evaluation protocol to quantify representational drift and recovery under permutations. From this protocol, we derive two scalar metrics:
    \begin{enumerate}
        \item $\mathrm{PI}_{\text{derange}}$: residual similarity under complete permutation (full derangement).
        \item $\rho_{\text{mono}}$: monotonicity of representational recovery as structural constraints are incrementally restored.
    \end{enumerate}

    \item \textbf{A prospective vision for permutation-invariant retrieval:} Using the above diagnostics and metrics as empirical grounding, we outline a research agenda toward \textbf{Platonic representations} for tables, meaning structure-aware embeddings designed to preserve PI. We discuss how such invariant-preserving encoders can serve as a foundation for table-centric retrieval and downstream reasoning, and identify open directions to make permutation-invariant retrieval a standard building block in IR pipelines.
\end{itemize}

\section{Background}
\label{sec:background}

\paragraph{\textbf{Why permutation invariance matters now}}
A central open question in table retrieval is whether table embeddings are robust to the structural symmetries of tables. A natural desideratum is \textit{PI}: semantically equivalent row/column permutations should yield similar embeddings. Yet this property has not been systematically characterized \emph{at the embedding level} for table representations, despite its value as a lightweight diagnostic for table-centric retrieval.

This gap is becoming increasingly consequential as the community moves toward LLM-powered interactive systems over structured data. Much of the work in Table Question Answering (TQA) \cite{li2024table}, TableGPT \cite{zha2023tablegpt}, and text-to-SQL \cite{10.14778/3641204.3641221} assumes that the relevant table (or database) is already provided. In prospective table-centric RAG settings, however, identifying the relevant table is itself the first bottleneck, and it depends directly on what the embedding model preserves about table semantics and structure. A related line of work evaluates table retrieval in end-to-end analytic query systems using embeddings from (non-)commercial LLMs \cite{ji2025target}, but does not explicitly test whether these embedders satisfy intrinsic table properties such as PI. One reason is the lack of clear, model-agnostic metrics for quantifying PI in table embeddings---a methodological gap that this paper aims to foreground.

\paragraph{\textbf{Limited diagnostics in table embedding space}}
Many structure-aware TRL approaches follow a largely \emph{task-oriented} paradigm: the impact of permutations is assessed indirectly through downstream performance (e.g., column type annotation, entity linking, relation extraction) rather than through direct analysis of the embedding space \cite{dash2022permutation}. Pre-trained table models such as TaBERT \cite{yin2020tabert}, TAPAS \cite{herzig2020tapas}, and TURL \cite{deng2020turl} incorporate table structure during pretraining, yet they are not explicitly designed to enforce PI. Similarly, strong tabular predictors such as TabNet \cite{arik2021tabnet} and FT-Transformer \cite{gorishniy2021revisiting} optimize task accuracy but do not target robustness properties of the learned representations.

More recently, HYTREL \cite{chen2023hytrel} introduced a fine-grained, cell-level assessment and showed that embeddings can be sensitive to row and column order. However, its evaluation primarily considered training-domain settings, limiting conclusions about out-of-distribution (zero-shot) robustness. Overall, a systematic and metric-agnostic assessment of PI \emph{in the embedding space} remains underexplored, despite its potential to guide the next generation of table retrievers.

\paragraph{\textbf{TRL bottleneck and the signal from benchmarks}}
Recent work on table-centric RAG increasingly explores hybrid architectures that combine dense retrieval with structured reasoning, e.g., BRIGHT \cite{su2025bright}, while TARGET \cite{ji2025target} emphasizes the need to evaluate the table retrieval component itself. In parallel, the evaluation landscape has evolved from simple cell-lookup tasks (e.g., WikiTableQA) toward reasoning-intensive benchmarks such as \textbf{BRIGHT} \cite{su2025bright} and \textbf{TableBench} \cite{wu2025tablebench}, which demand logically grounded retrieval and industrial-scale multi-step reasoning. Despite this progress, these benchmarks expose a persistent representation bottleneck: most pipelines still rely on \textit{table linearization} to fit tables into standard language-model inputs, and even instruction-tuned approaches such as \textit{TableLlama} \cite{zhang2024tablellama} do not reliably preserve table structure in their embeddings. As we will demonstrate later, this manifests as \textit{structural perturbation}---a failure to achieve true row/column PI---and it amplifies \textit{linearization bias}, where retrieval defaults to localized text matching rather than global relational and quantitative reasoning. Collectively, these observations motivate our diagnostic focus and prospective agenda: table representation learning must explicitly target structural invariants as a foundation for permutation-invariant retrieval in future table-RAG systems.

\section{Methodology: Diagnosing Serialization Bias in TRL}
\label{sec:method}

We diagnose serialization bias in LLM-based table representations by comparing embeddings from linearized tables with those produced by a table-structure-aware encoder. Building on this setup, we instantiate the PRH for tables and formalize how PI can be assessed in the embedding space as an intrinsic diagnostic of this bias. To make this comparison easier to follow, we first present the tables and introduce our notation.

Let a table \(T\) have \(n\) rows and \(m\) columns with cell set
\[C=\{c_{ij}\}_{i\in[n],\,j\in[m],}\] row headers \(R=\{r_i\}_{i\in[n]}\), and column headers
\(H=\{h_j\}_{j\in[m]}\). We write \(T=C\cup R\cup H\) with \(|C|=nm\).

Let \(T^{\ast}\) denote the space of unstructured text (strings). We convert any (semi-)structured table into an LLM-readable sequence via a linearization function
\(
\delta:\ T \longrightarrow T^{\ast},
\)
e.g., row-wise serialization with special delimiters for \texttt{[ROW]}, \texttt{[CELL]}, and \texttt{[SEP]} to preserve extractability of each cell span within the sequence as implemented in RAG frameworks.

\subsection{Serialization: LLM-based table cell embeddings}\label{subsec:LLM-based}
Given a serialized table \(\delta(T)\), we obtain a contextual embedding for any cell
\(x\in T\) using a base embedding function
\begin{equation}
\phi:\ (x,\delta(T)) \longmapsto \phi(x \mid \delta(T)) \in \mathbb{R}^{d},
\label{eq:LLM_embedding_function}
\end{equation}
where \(d\) is the embedding dimension and \(\phi\) depends on the LLM family
(encoder-only models use bidirectional context; decoder-only models use causal context).
For simplicity, we consider that the LLM-only table embedding space is produced by the function $\Phi$:
\[
\Phi(T)
=\big[\phi(c_{11}\!\mid\!\delta(T)),\ldots,\phi(c_{nm}\!\mid\!\delta(T))\big]^{\top}
\in \mathbb{R}^{nm\times d}.
\]

Because \(\phi(\cdot\mid\delta(T))\) conditions on the serialized table, identical surface values can receive different embeddings when their headers differ. For example, the numeric value \(\texttt{12}\) under the column header \emph{Age} may be embedded differently from \(\texttt{12}\) under \emph{Weight}, reflecting context capturing when LLMs are applied to produce table cell embeddings.

\subsection{Structure-aware cell embeddings via TRL}
We compare \emph{LLM linearization} to a \emph{structure–aware} TRL embedder based on
\emph{Semantic Meta-Paths} (SMP)~\cite{TCHUITCHEU2024110734}, which help to preserve header–cell semantic context:
\begin{equation}
T \xrightarrow{\ \phi\ }\ D_y \xrightarrow{\ \mathrm{SMP}\ }\ S
\ \xrightarrow[\ \psi_{\theta}\ ]{\ \mathrm{TRL}\ }\ D_z.
\label{eq:TRL_pipeline}
\end{equation}
where \(D_y=\{\phi(x):x\in T\}\) are the initial context-free LLM-based cell embeddings and and \(D_z=\{\psi_{\theta}(x\!\mid\! S):x\in T\}\) are \emph{context-aware} embeddings produced by the TRL encoder \(\psi_{\theta}\) ($\theta$ denotes trainable parameters of the encoder) given the set \(S\) of SMP sequences extracted from \(T\). SMPs extract \emph{semantically meaningful paths} (e.g., header \(\leftrightarrow\) cell \(\leftrightarrow\) header), mimicking human table reading by actively tying each cell to its row/column headers. TRL optimizes cell embeddings based on two complementary signals: (i) \emph{local alignment (intra-SMP):} pull together a cell and its corresponding headers, enforcing header–cell correspondence within each SMP; and
(ii) \emph{global separation (inter-SMP):} push apart embeddings of semantically distinct cells (even if their surface text is similar), increasing the distance across SMPs' embeddings. Combining these two signals, TRL aims to learn cell embeddings by linking them to the corresponding headers. We refer to the TRL-based table embedding space as
\[
\Psi_{\theta}(T)=\big[\psi_{\theta}(c_{11}\!\mid\! S),\ldots,\psi_{\theta}(c_{nm}\!\mid\! S)\big]^{\top}
\in\mathbb{R}^{nm\times d}.
\]

\subsection{Platonic Representation Hypothesis (PRH) in TRL}
\label{subsec:platonic}

Let \( G = S_n \times S_m \) be the product of symmetric groups acting on row and column indices of a table \( T \). We define a group action \( h: G \times T \to T \) such that for any \((\sigma, \tau) \in G\) and table element \( x \in T \):
\begin{equation}
 h{(\sigma,\tau)}(x)=
\begin{cases}
c_{\sigma(i)\,\tau(j)} & \text{if } x=c_{ij},\\
r_{\sigma(i)}          & \text{if } x=r_i,\\
h_{\tau(j)}            & \text{if } x=h_j.
\end{cases}  \label{goup_action} 
\end{equation}

Since \(\sigma \in S_n,\tau \in S_m\) are bijections, \(h_{(\sigma,\tau)}\) is a well-defined group action. We investigate the PRH within tables by quantifying the invariance of cell embeddings under the group action $h$. For any \(g\in G\), we compare embeddings computed on the permuted table \(g\!\cdot\!T:=h_g(T)\) to those on \(T\) using a similarity score \(s(\cdot,\cdot)\):
\begin{align}
\text{(Intra-model)}\quad
& s\!\big(\Psi_{\theta}(g\!\cdot\! T),\, \Psi_{\theta}(T)\big), \label{eq:platonic_intra}\\
\text{(Cross-model)}\quad
& s\!\big(\Psi_{\theta}(g\!\cdot\! T),\, \Psi_{\beta}(T)\big), \label{eq:platonic_cross}\\
\text{(LLM baseline)}\quad
& s\!\big(\Phi(g\!\cdot\! T),\, \Phi(T)\big). \label{eq:platonic_llm}
\end{align}
Here, \(\Psi_{\theta}\) and \(\Psi_{\beta}\) denote TRL encoders with different parameters, and \(\Phi\) represents the pretrained LLM-derived cell embeddings function, as explained in Section \ref{subsec:LLM-based}. Higher scores across permutations \(g \in G\) indicate stronger PI, thereby providing more substantial evidence for the PRH for tables.


\label{subsec:metric}
To compare embedding spaces (i.e., Equations. \ref{eq:platonic_intra}, \ref{eq:platonic_llm}), we employ \emph{Centered Kernel Alignment (CKA)}~\cite{kornblith2019similarity},
a normalized HSIC-based similarity widely used for cross-network representation analysis
(related tools include CCA/SVCCA~\cite{morcos2018insights,raghu2017svcca}):
\begin{equation}
\mathrm{CKA}(X,Y)=
\frac{\mathrm{HSIC}(X,Y)}{\sqrt{\mathrm{HSIC}(X,X)\,\mathrm{HSIC}(Y,Y)}},
\label{eq:metric}
\end{equation}
where \(X\) and \(Y\) are the cell-embeddings of the original table \(T\) and its permuted counterpart \(g\!\cdot\! T\), respectively. This formulation includes several desirable properties, such as invariance to: 
\begin{itemize}
    \item Isotropic Scaling, i.e., \(\text{CKA}(X,Y) = \text{CKA}(\alpha_1 X, \alpha_2 Y)\) for any \(\alpha_1, \alpha_2 \in \mathbb{R}^+\).
    \item Orthogonal Transformation, i.e., \( \allowdisplaybreaks \text{CKA}(X,Y) = \text{CKA}\\(XU, YV)\) for any full-rank orthogonal matrices \(U\) and \(V\) such that \(U^TU = I\) and \(V^TV = I\).
\end{itemize}
These properties make CKA a well-suited choice for studying the PI of \(\Phi(\cdot)\), \(\Psi_{\theta}(\cdot)\), and \(\Psi_{\beta}(\cdot)\).


\begin{figure*}[t]
\centering
\centering
\resizebox{\linewidth}{!}{%
\begin{tabular}{@{}lccccccccccc@{}}
\toprule
\textbf{Club} & \textbf{Played} & \textbf{Won} & \textbf{Drawn} & \textbf{Lost} &
\textbf{Points for} & \textbf{Points against} & \textbf{Tries for} & \textbf{Tries against} &
\textbf{Try bonus} & \textbf{Losing bonus} & \textbf{Points} \\
\midrule
Bridgend Athletic RFC & 22 & 16 & 0 & 6 & 523 & 303 & 68 & 31 & 10 & 4 & 78 \\
Builth Wells RFC      & 22 & 17 & 0 & 5 & 473 & 305 & 57 & 29 &  7 & 2 & 77 \\
Kidwelly RFC          & 22 & 14 & 1 & 7 & 532 & 386 & 63 & 45 &  5 & 3 & 66 \\
Loughor RFC           & 22 & 13 & 1 & 8 & 532 & 388 & 69 & 43 &  9 & 1 & 64 \\
Ammanford RFC         & 22 & 13 & 0 & 9 & 447 & 394 & 58 & 51 &  6 & 4 & 62 \\
Waunarlwydd RFC       & 22 & 12 & 2 & 8 & 504 & 439 & 57 & 55 &  6 & 3 & 61 \\
Pencoed RFC           & 22 & 13 & 0 & 9 & 425 & 328 & 53 & 36 &  4 & 4 & 60 \\
BP RFC                & 22 &  9 & 1 &12 & 367 & 358 & 39 & 43 &  2 & 7 & 47 \\
Mumbles RFC           & 22 &  8 & 2 &12 & 373 & 450 & 50 & 56 &  4 & 4 & 44 \\
Cwmavon RFC           & 22 &  6 & 2 &14 & 332 & 515 & 39 & 66 &  3 & 5 & 36 \\
Penclawdd RFC         & 22 &  4 & 1 &17 & 263 & 520 & 28 & 68 &  1 & 3 & 22 \\
Gorseinon RFC         & 22 &  2 & 0 &20 & 340 & 725 & 48 &106 &  3 & 4 & 15 \\
\bottomrule
\end{tabular}}

\caption{Single-table test example of Rugby Club Performance Statistics ( from WikiSQL dataset, Table ID 560). Identical content values (e.g., ``22'') can denote different semantics depending on headers.}
\label{tab:rugby_stats}
\end{figure*}

\begin{figure*}[h]
\begin{minipage}[b]{.31\linewidth}
        \centering
        \includegraphics[height=2.5cm, width=\linewidth]{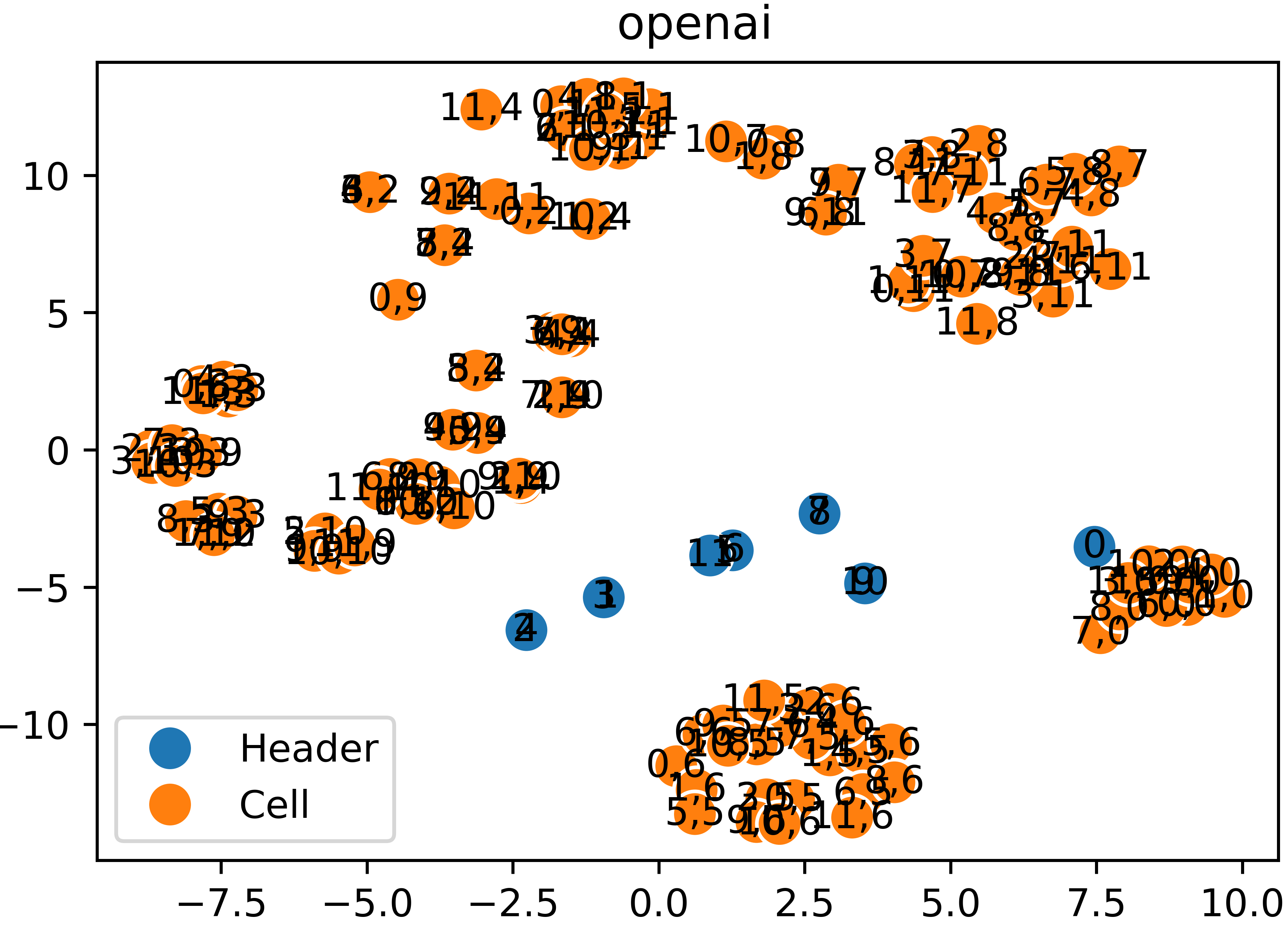} 
    \end{minipage}\hspace*{-10em}
    \hfill 
    \begin{minipage}[b]{0.31\textwidth}
        \centering
        \includegraphics[height=2.5cm,width=\linewidth]{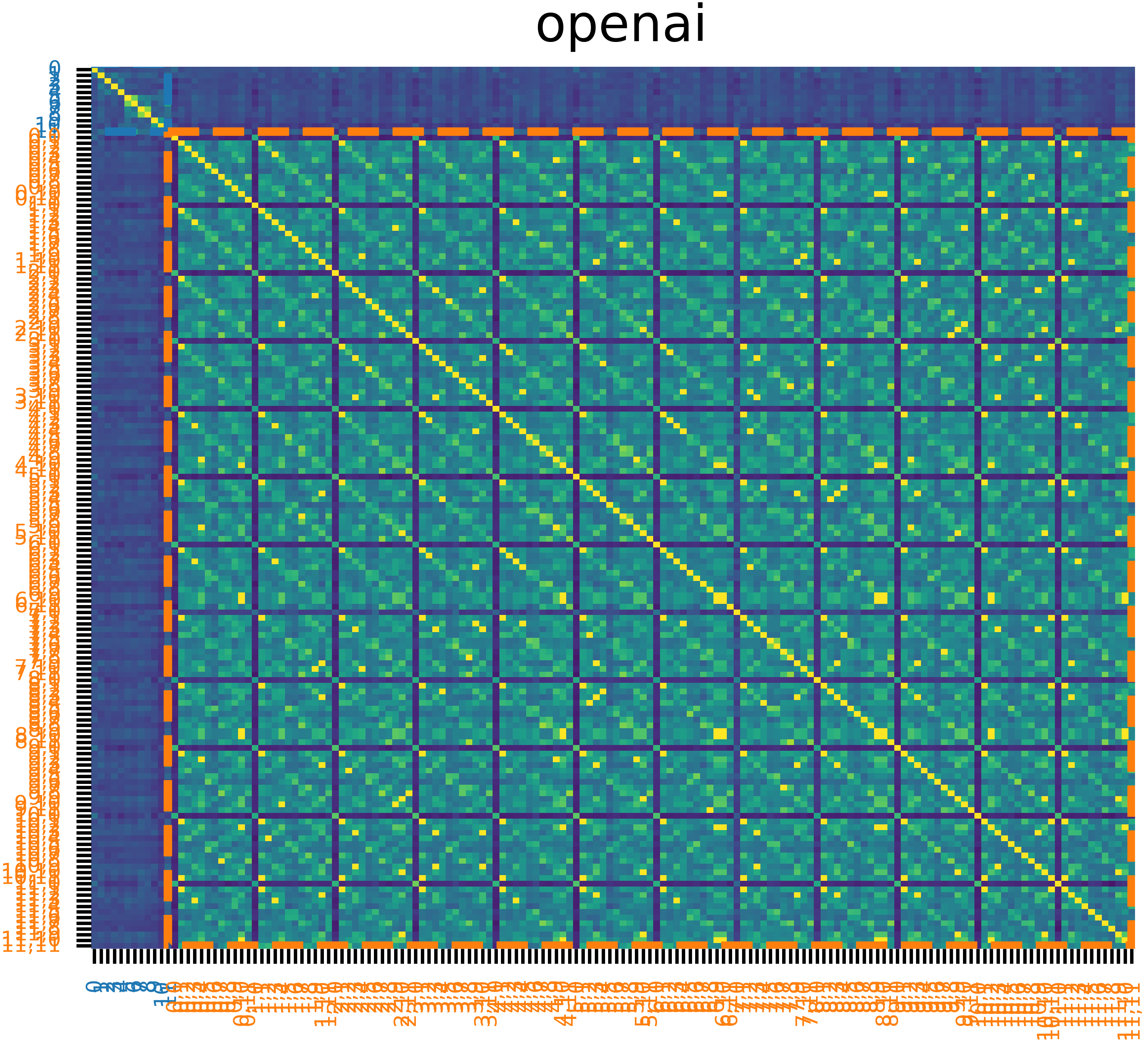} 
    \end{minipage}\hspace*{-10em}
    \hfill 
    \begin{minipage}[b]{.31\linewidth}
        \centering
        \includegraphics[height=2.5cm,width=\linewidth]{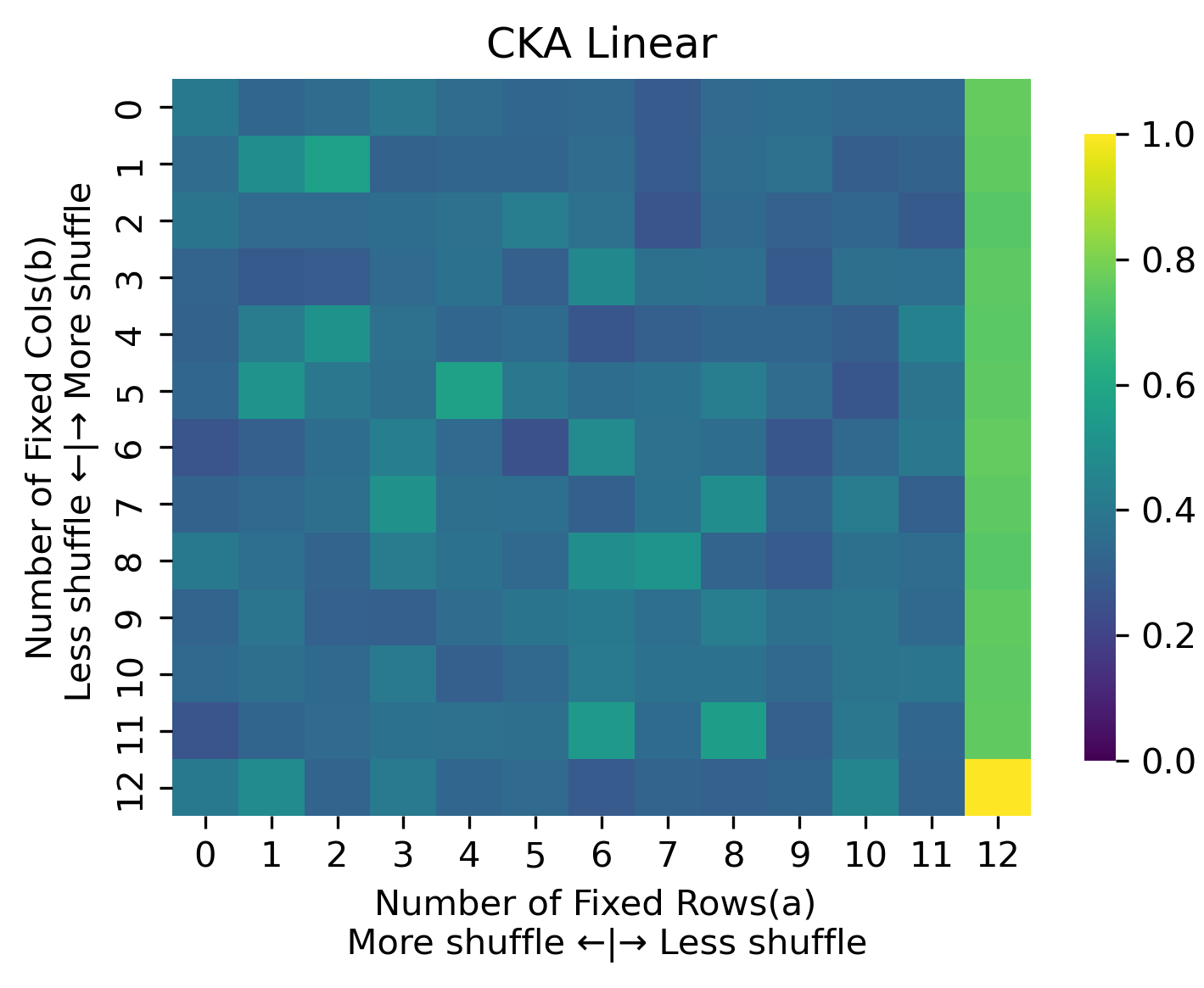} 
    \end{minipage}
      \vfill
    \begin{minipage}[b]{.31\linewidth}
        \centering
        \includegraphics[height=2.5cm, width=\linewidth]{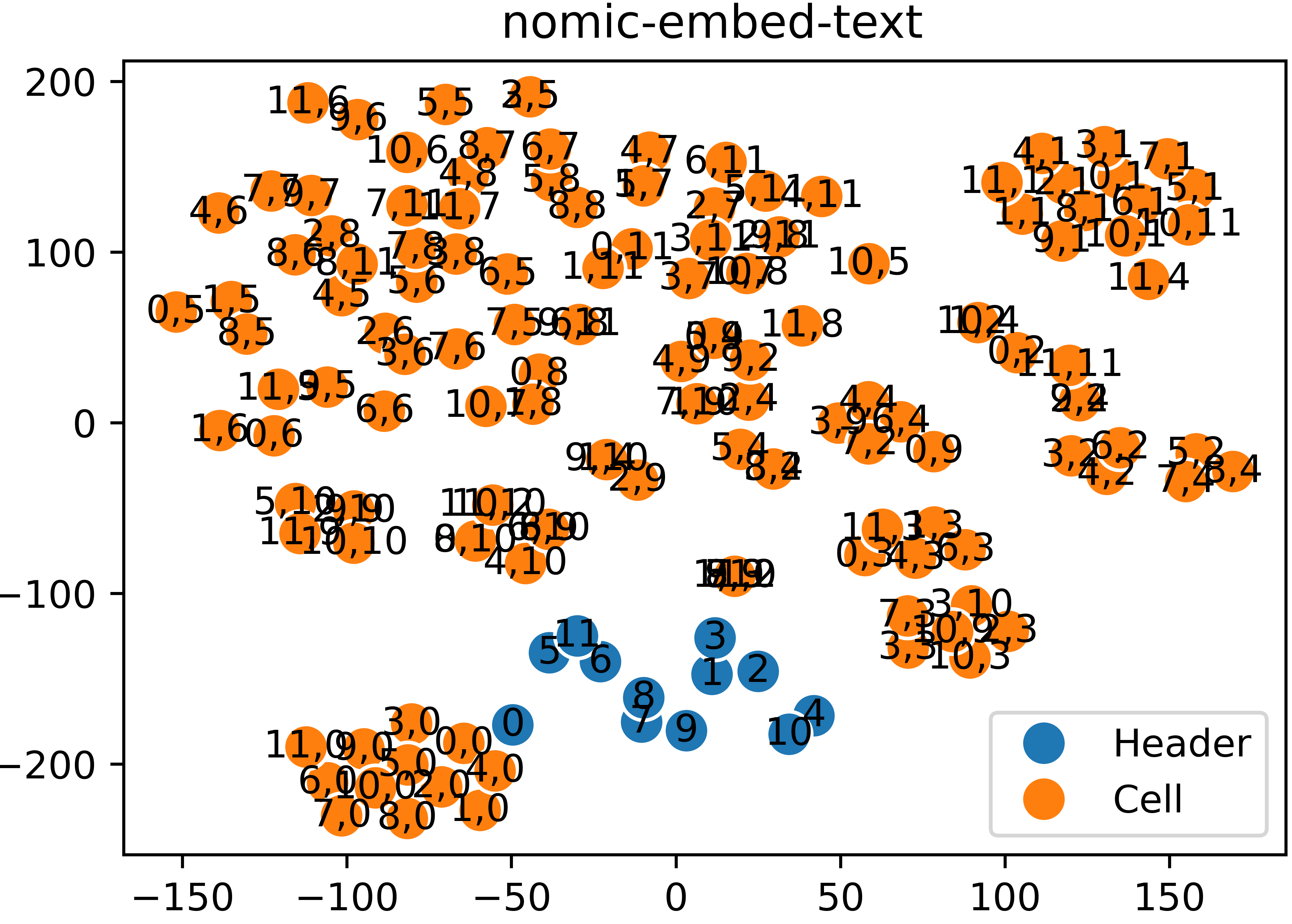} 
    \end{minipage}\hspace*{-10em}
    \hfill 
    \begin{minipage}[b]{0.31\textwidth}
        \centering
        \includegraphics[height=2.5cm,width=\linewidth]{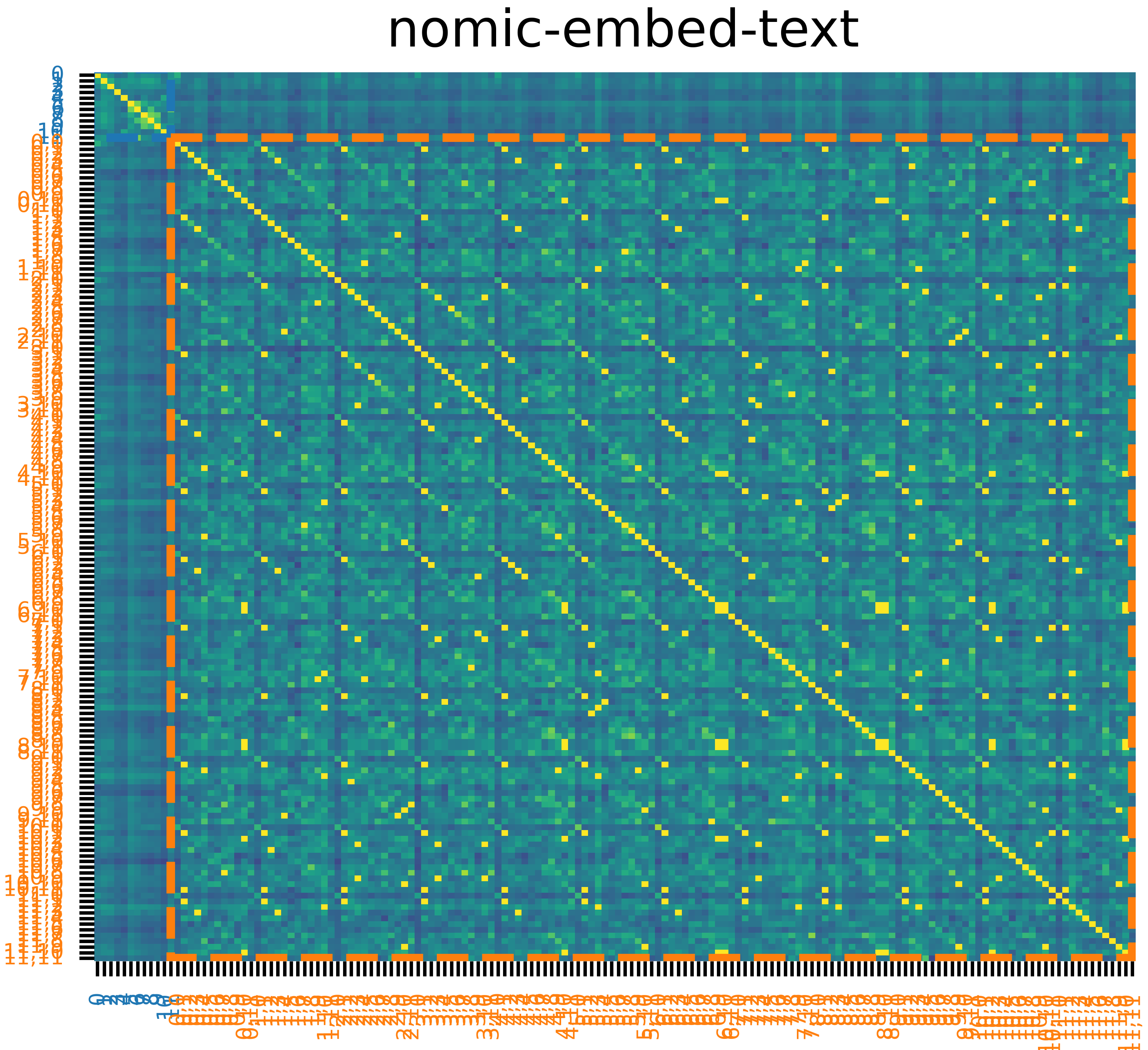} 
    \end{minipage}\hspace*{-10em}
    \hfill 
    \begin{minipage}[b]{.31\linewidth}
        \centering
        \includegraphics[height=2.5cm,width=\linewidth]{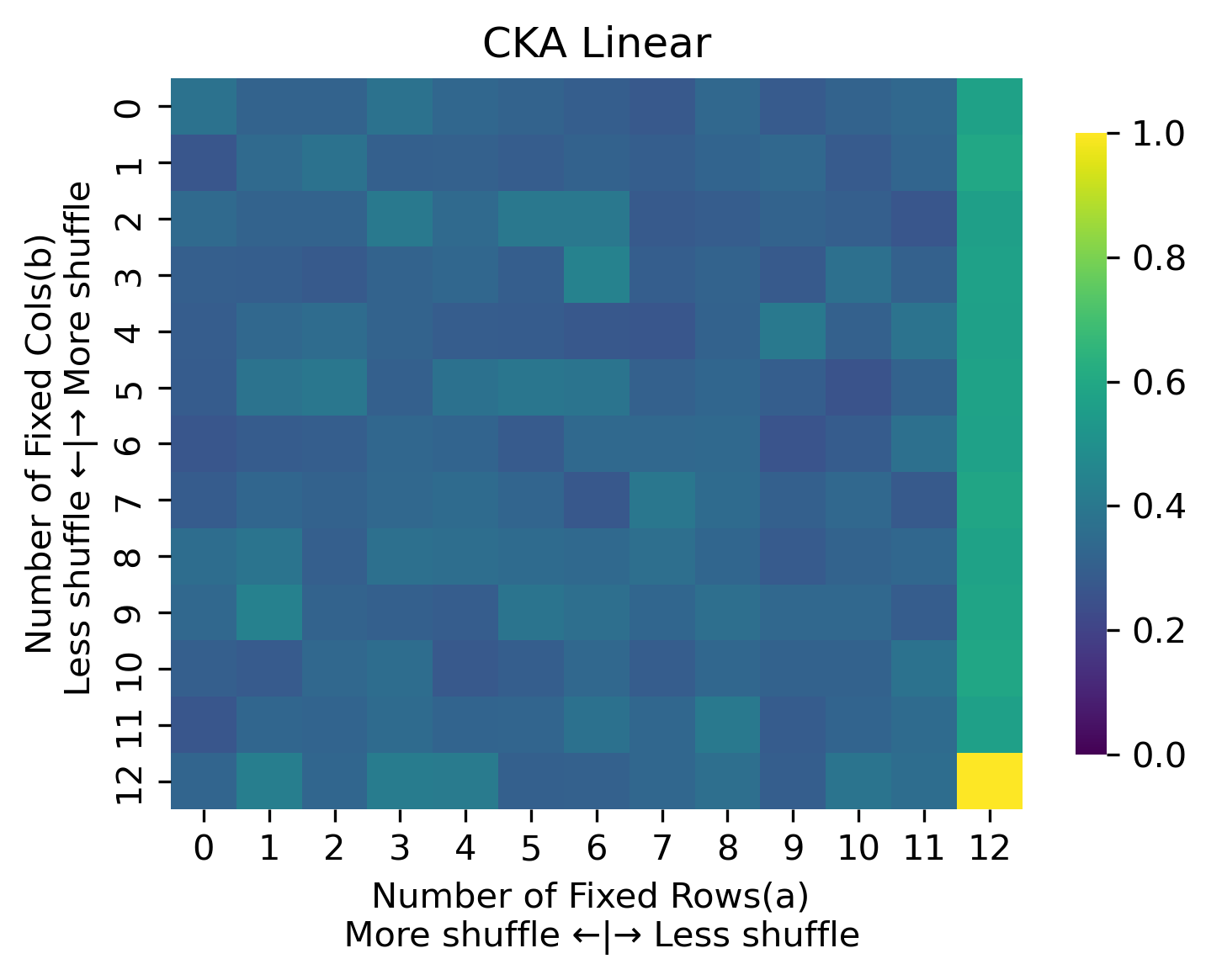} 
    \end{minipage}
    \vfill
    \begin{minipage}[b]{.31\linewidth}
        \centering
        \includegraphics[height=2.5cm, width=\linewidth]{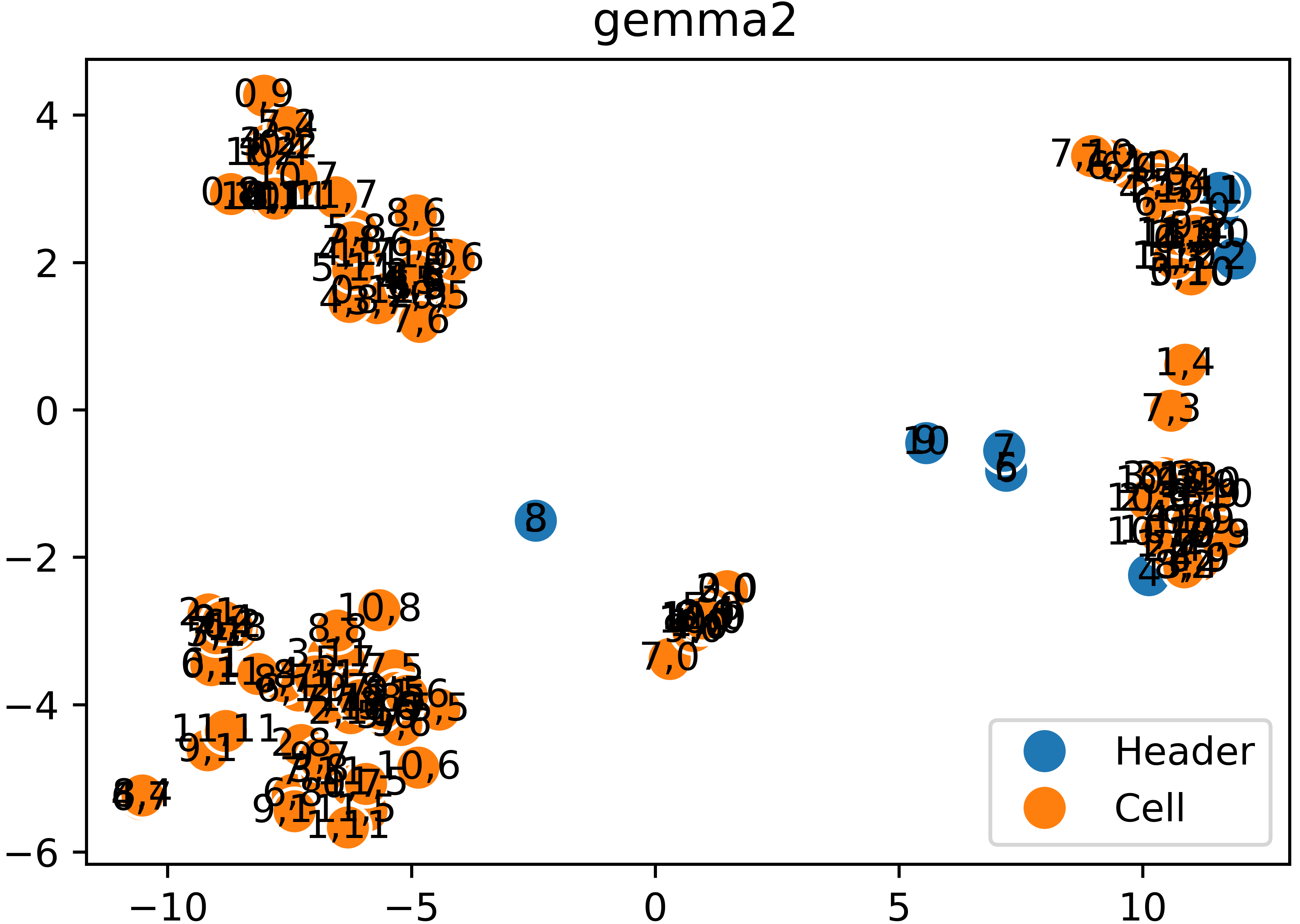} 
    \end{minipage}\hspace*{-10em}
    \hfill 
    \begin{minipage}[b]{0.31\textwidth}
        \centering
        \includegraphics[height=2.5cm,width=\linewidth]{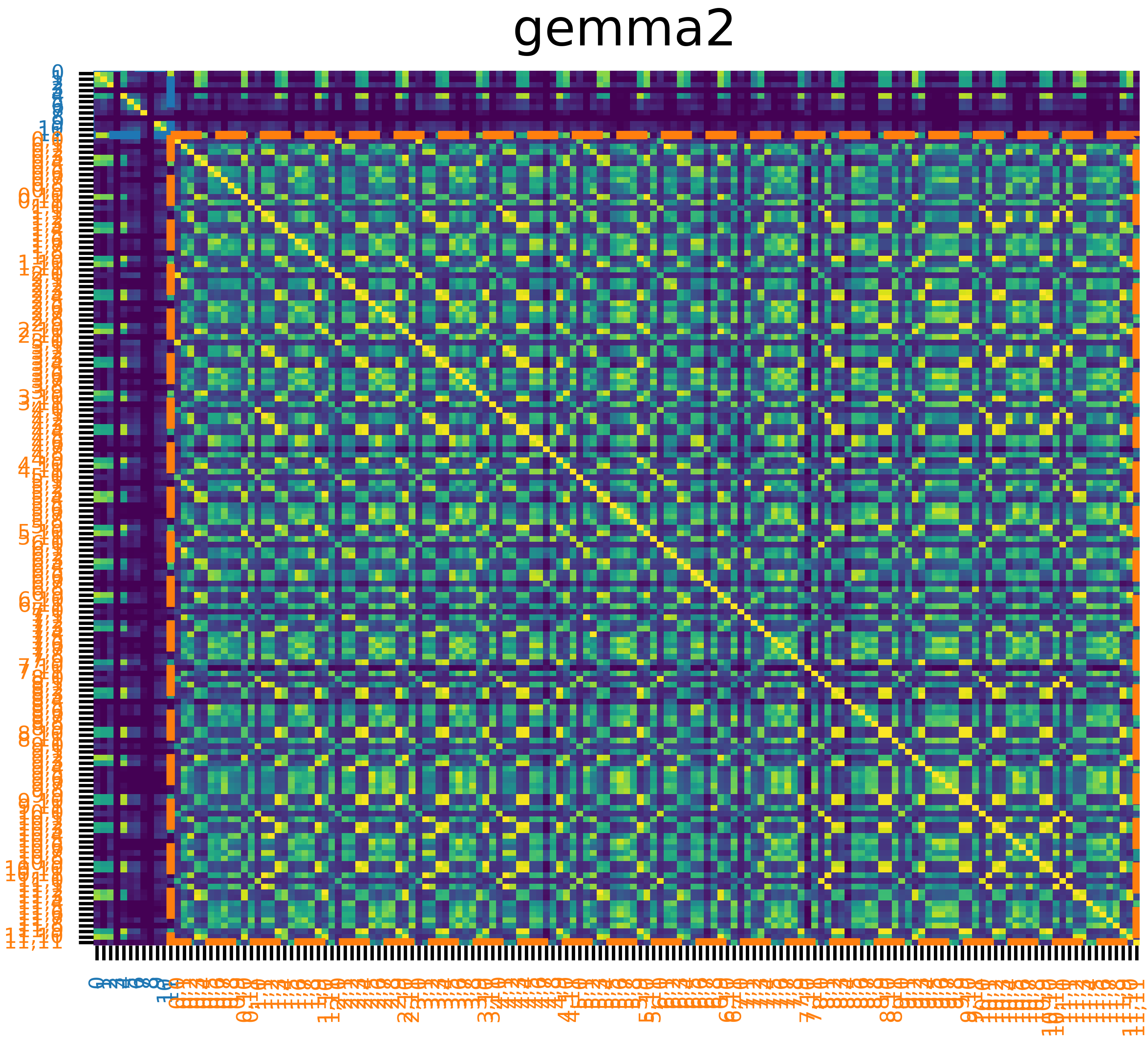} 
    \end{minipage}\hspace*{-10em}
    \hfill 
    \begin{minipage}[b]{.31\linewidth}
        \centering
        \includegraphics[height=2.5cm,width=\linewidth]{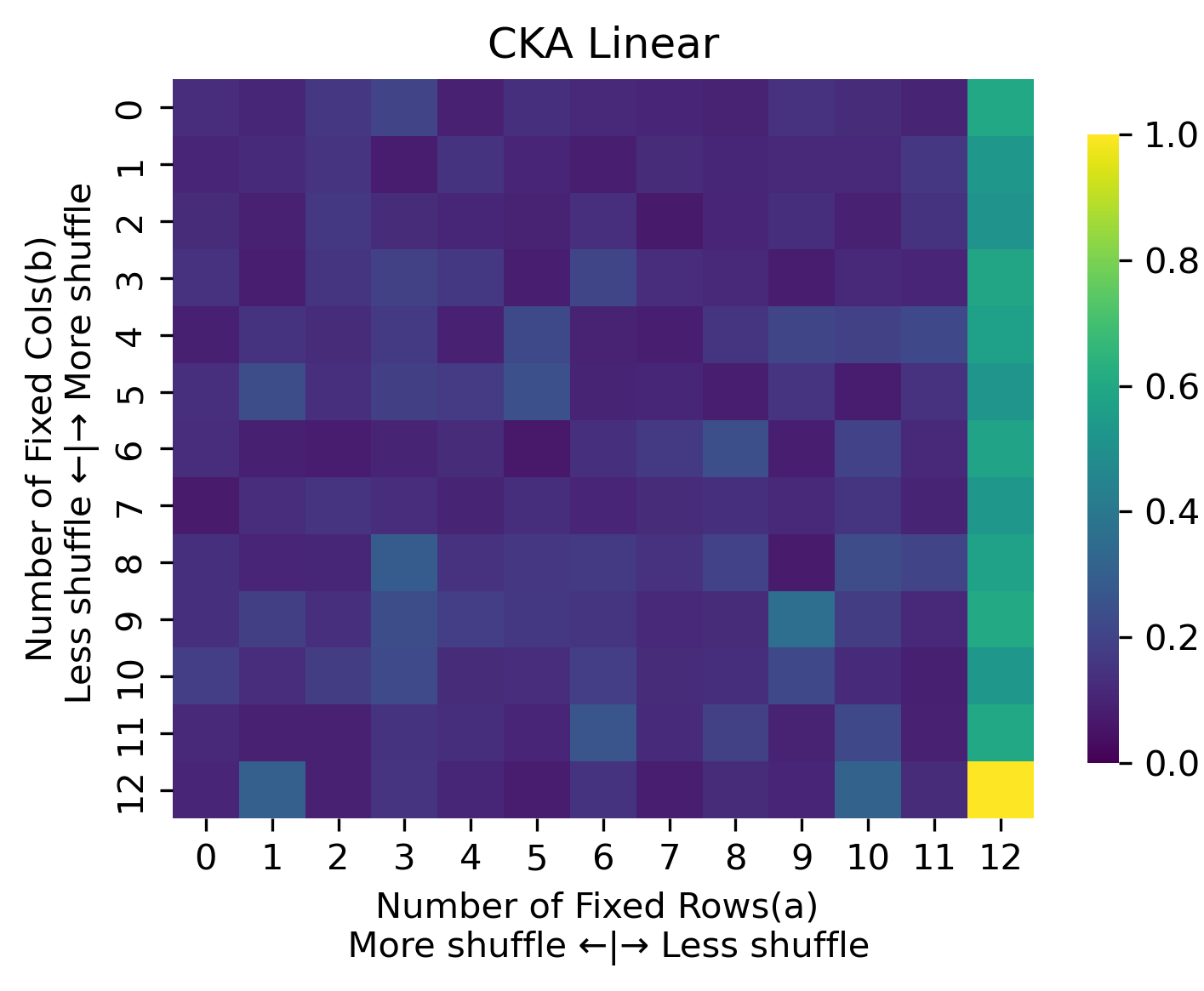} 
    \end{minipage}
    
    \vfill
    \begin{minipage}[b]{.31\linewidth}
        \centering
        \includegraphics[height=2.5cm, width=\linewidth]{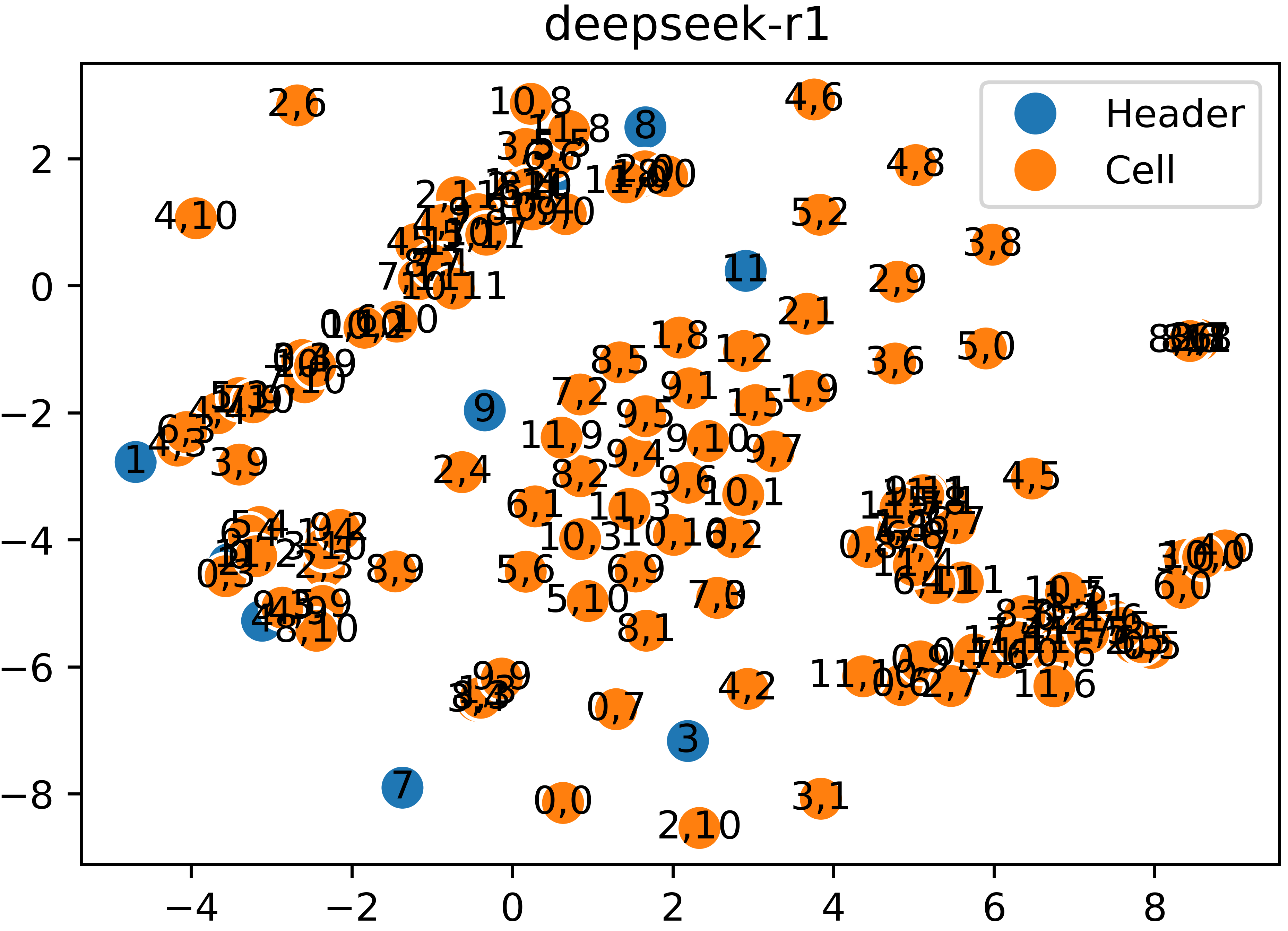} 
    \end{minipage}\hspace*{-10em}
    \hfill 
    \begin{minipage}[b]{0.31\textwidth}
        \centering
        \includegraphics[height=2.5cm,width=\linewidth]{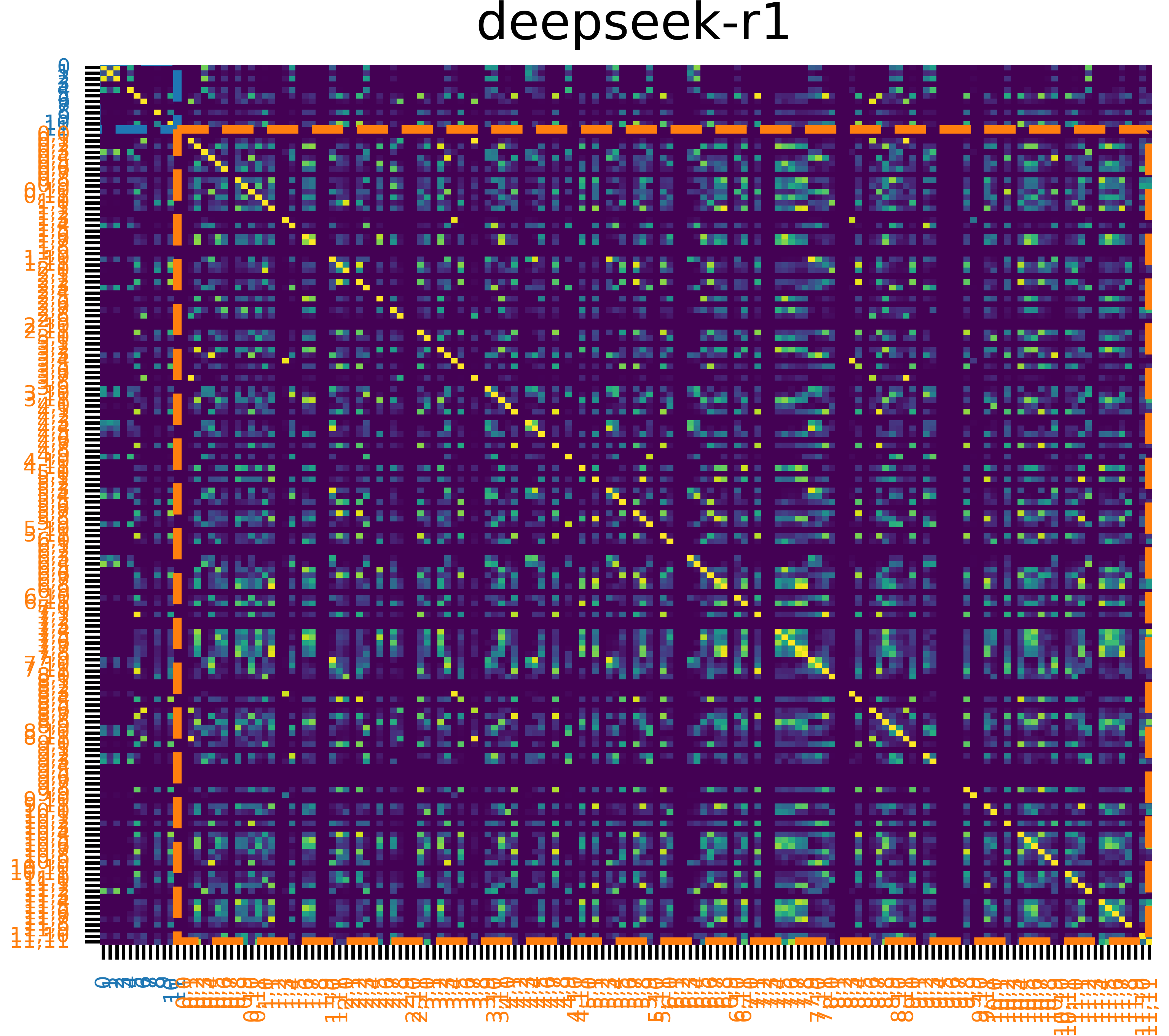} 
    \end{minipage}\hspace*{-10em}
    \hfill 
    \begin{minipage}[b]{.31\linewidth}
        \centering
        \includegraphics[height=2.5cm,width=\linewidth]{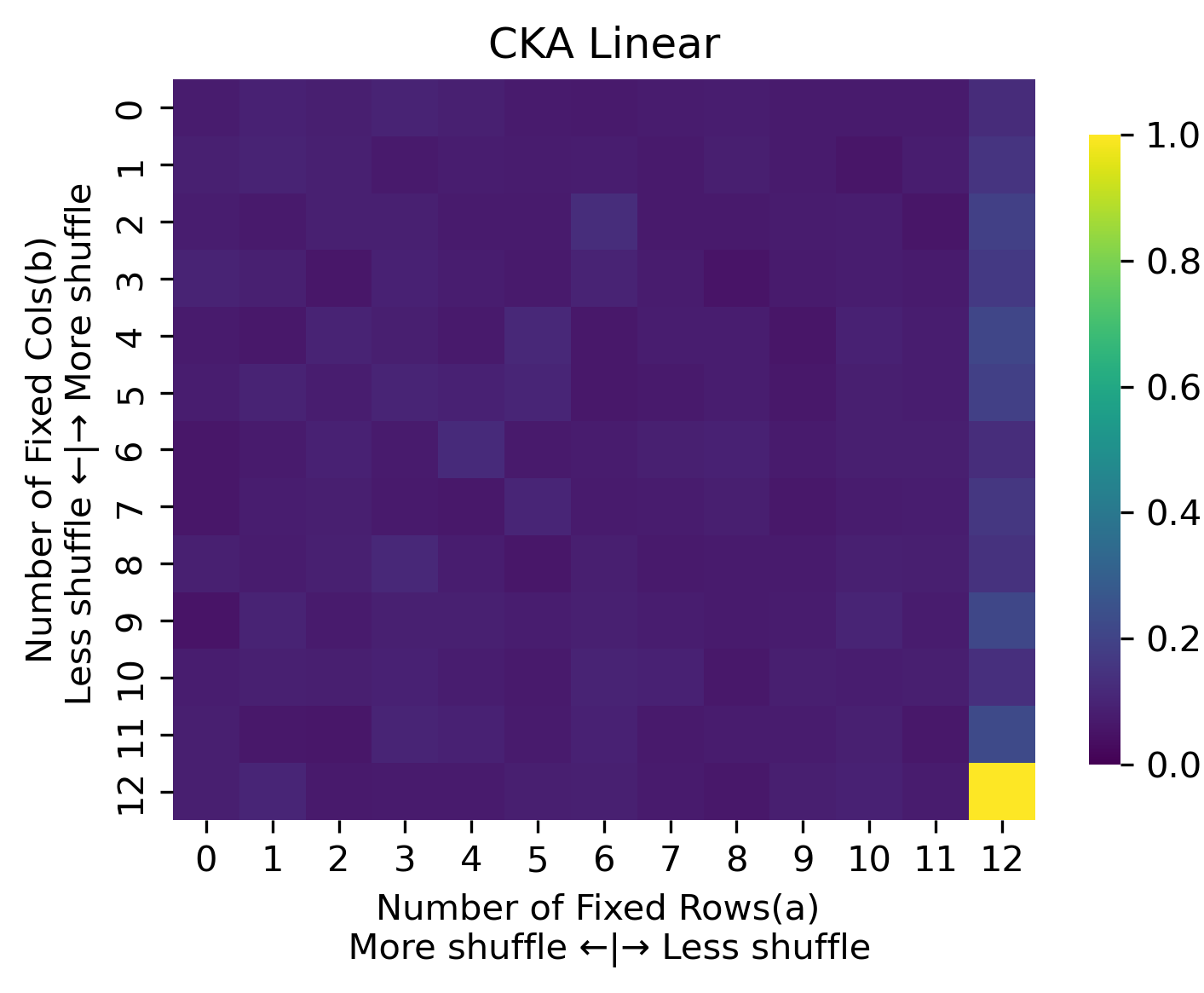} 
    \end{minipage}
    \vfill
    \begin{minipage}[b]{.31\linewidth}
        \centering
        \includegraphics[height=2.5cm, width=\linewidth]{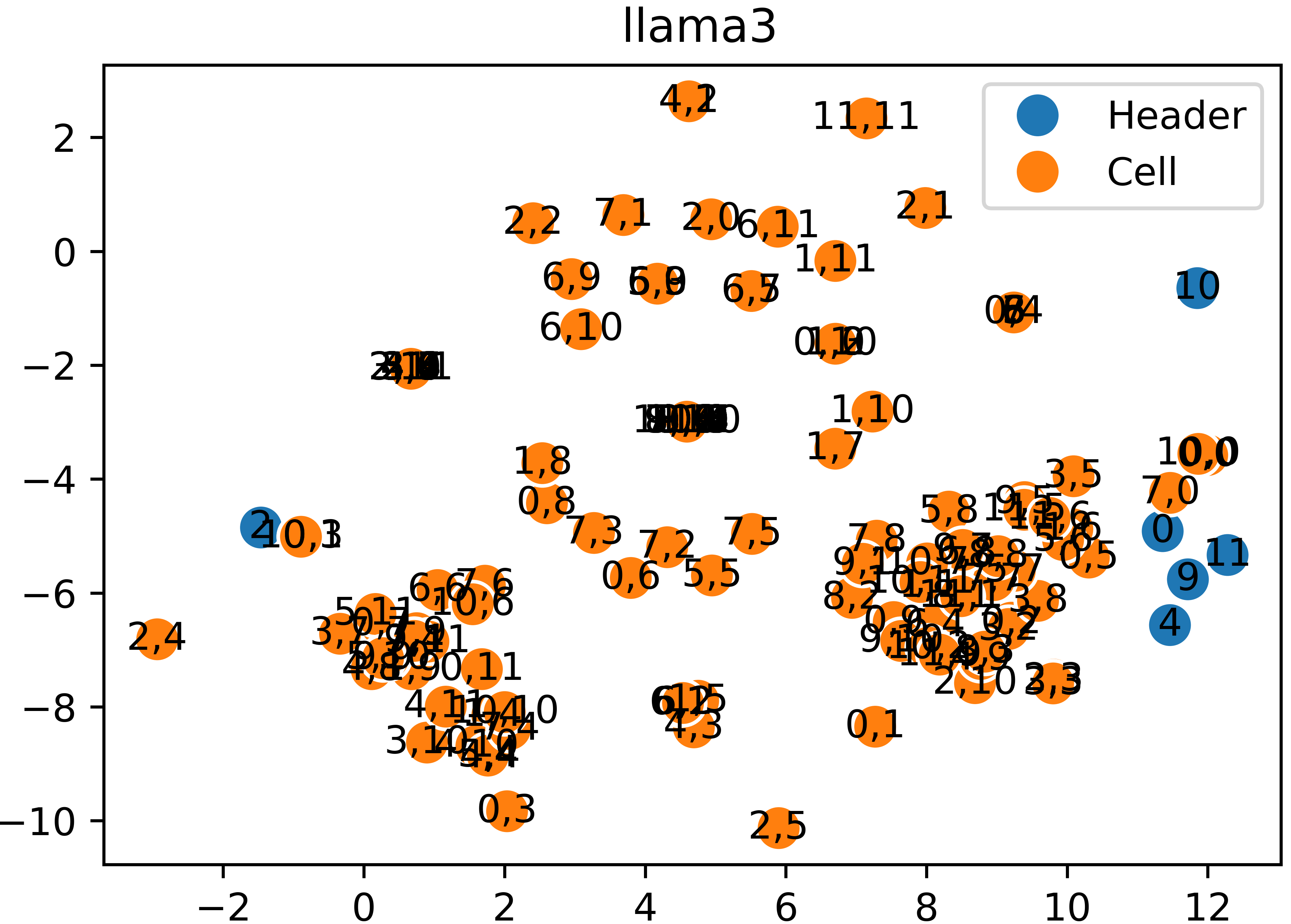} 
    \end{minipage}\hspace*{-10em}
    \hfill 
    \begin{subfigure}[b]{0.31\textwidth}
        \centering
        \includegraphics[height=2.5cm,width=\linewidth]{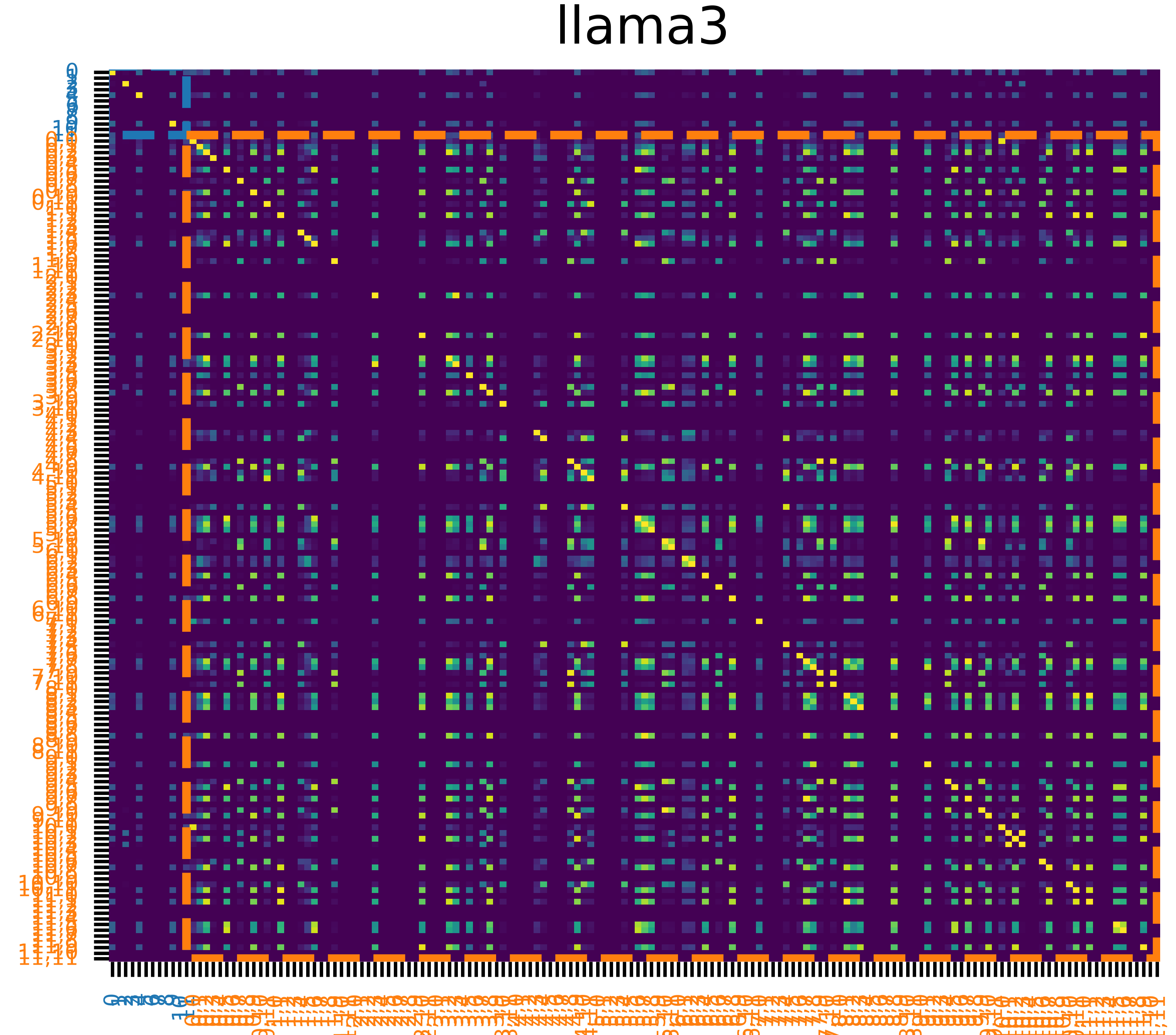} 
    \end{subfigure}\hspace*{-10em}
    \hfill 
    \begin{minipage}[b]{.31\linewidth}
        \centering
        \includegraphics[height=2.5cm,width=\linewidth]{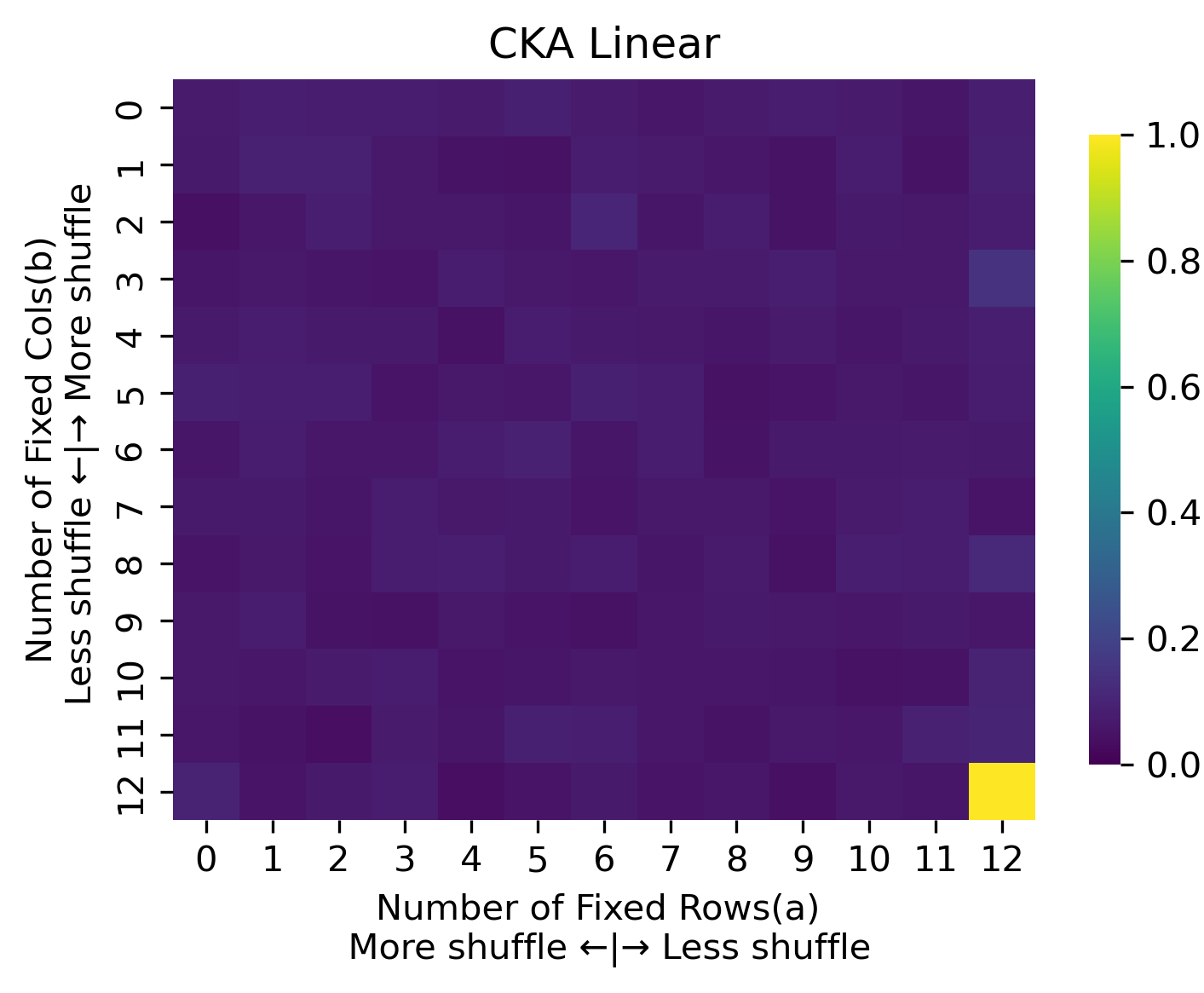} 
    \end{minipage}
 \caption{LLM-only table embeddings across LLMs (\texttt {{OpenAI}}, \texttt{nomic-embed-text}, \texttt{Gemma2}, \texttt{DeepSeek-R1}, and \texttt{llama3}). 
Rows correspond to models. Columns show: \textbf{Left} t\mbox{-}SNE projections of LLM-based \emph{cell embeddings} for the baseline table (Figure~\ref{tab:rugby_stats}); a point indexed by $(\mathrm{row\_id},\mathrm{col\_id})$ denotes the projected embedding of a cell. 
\textbf{Center} pairwise cosine similarity within the resulting cell-embedding set. 
\textbf{Right} heatmaps $H[a,b]$ of similarity between the original table and its $169$ permutations $T_a^b$.}

    \label{fig:LLM-based embedding}
\end{figure*}

\begin{figure*}[t]
    \centering
    \includegraphics[height=10cm, width=0.95\linewidth]{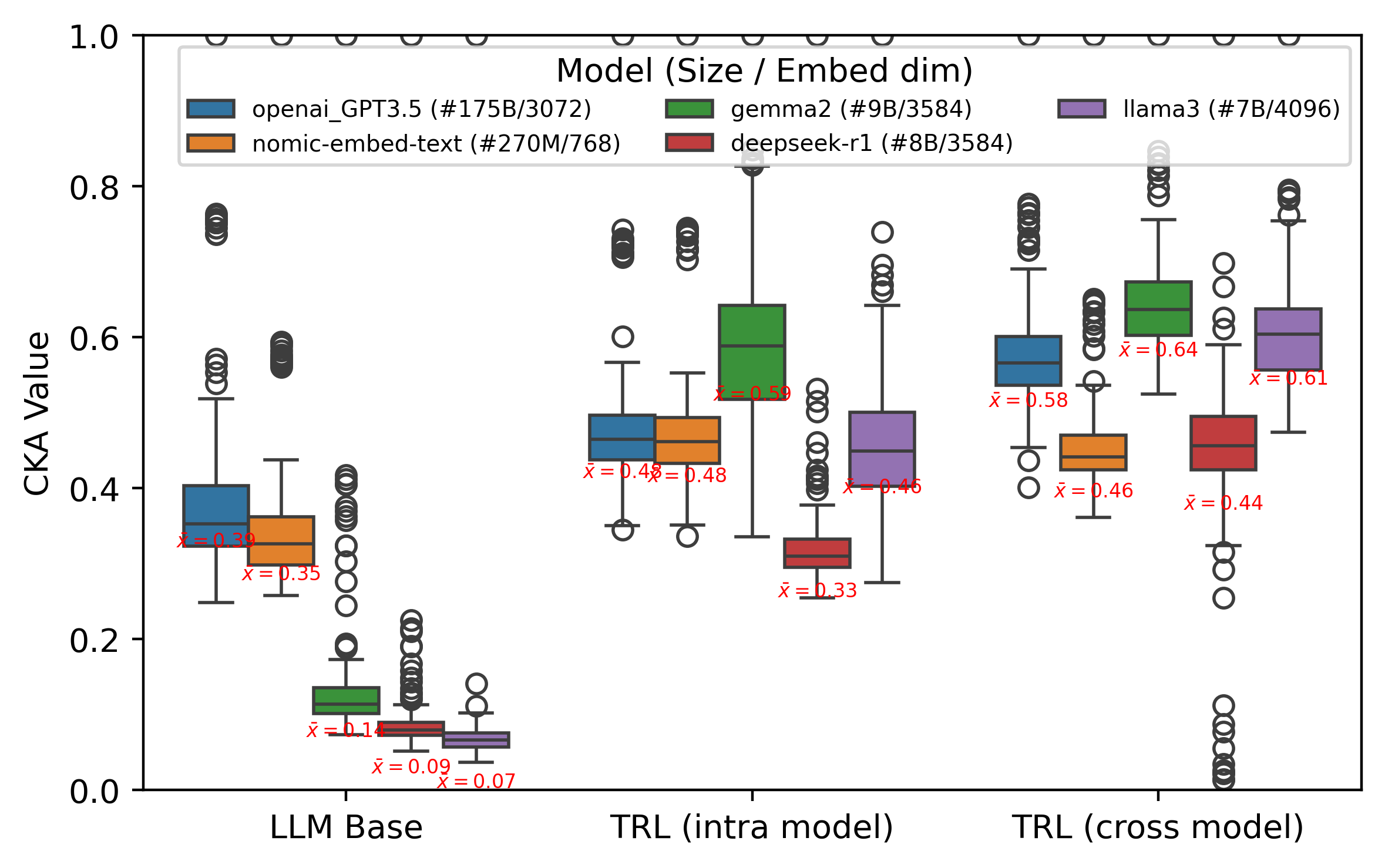}
    \caption{Distribution of heatmap entries \(H[a,b]\) and the mean $\Bar{x}$ comparing LLM and TRL-derived cell embeddings under two settings: (i) \emph{intra-model}, which examines embeddings of permuted tables generated by the same model that was trained on the original table.
    (ii) \emph{Cross-model}, where embeddings of permuted tables are extracted from different models trained on each other’s samples.}
    \label{fig:cka-boxplot_LLM-embed}
\end{figure*}

\section{Experiments and Discussion}
\label{sec:experiments}

To probe \emph{Platonic} invariance from a table \(T\) with \(n\) rows and \(m\) columns, we use the row–column permutation action $h$ (see Equation \ref{goup_action}) to build a grid of controlled permutations parameterized by the number of fixed rows and columns $a$ and $b$. Meaning choosing
\(\sigma_a \in S_n\) and \(\tau_b \in S_m\) such that \(a=\mathrm{fix}(\sigma_a)\!=\!|\{i:\,\sigma_a(i)=i\}|\), and \(b=\mathrm{fix}(\tau_b)\!=\!|\{j:\,\tau_b(j)=j\}|\). We refer to a parameterized $(a,b)$-permuted table as \(T_a^b :=h\big((\sigma_a,\tau_b),\!T\big):=h_g(T)\), yielding a test bench
\[
\{\,T_a^b:\ a=0, \cdots, n;\ b=0, \cdots,m\},
\]
which contains \((n+1)(m+1)\) permuted tables, spanning identity (i.e., $T_n^m$) to full derangement $T_0^0$ (maximal shuffling on both rows and columns).

\subsection{Proposed metrics and models baseline}
\noindent \textbf{Proposed metrics.} For each \((a,b)\), we compare the cell embedding matrix of the original table
\(X=\Psi_\theta(T)\) (or \(\Phi(T)\) for LLM baselines) with that of the permuted table
\(Y_a^b=\Psi_\theta(T_a^b)\) (or \(\Phi(T_a^b)\)), producing a \emph{Platonic heatmap} $H \in [0,1]^{(n+1)\times(m+1)}$ such that:
\begin{equation}
H[a,b] =\mathrm{CKA}\!\big(X,\ Y_a^b\big), \forall (a,b) \in \{0,\dots,n\}\times\{0,\dots,m\}
\label{eq:heatmap}
\end{equation}
This heatmap is further summarised with two scalars:
\begin{equation}
    \mathrm{PI}_{\text{derange}} =  H[0,0], \quad
\rho_{\text{mono}} = \operatorname{Spearman}\!\big(H[a,b],\, a{+}b\big). \label{eq:pi-mono}
\end{equation}

Here, \(\mathrm{PI}_{\text{derange}}\) quantifies Permutation Invariance ($\mathrm{PI}$) under \emph{full derangement} \(T_0^0\) (higher is better), and \(\rho_{\text{mono}}\in[-1,1]\) measures $H[a,b]$ changes as structure is restored $T_n^m$ (i.e., as \(a\) and \(b\) increase toward \(n\) and \(m\)): values near \(+1\) indicate strong increasing
monotonicity from $\mathrm{PI}_{\text{derange}}$, near \(0\) no monotonic trend, and near \(-1\) strong decreasing monotonicity.

As an additional summary, we also report a normalized Area Under Curve (AUC) computed along a single axis of restoration, either rows (\(a\)) or columns (\(b\)), while keeping the other axis fixed. Formally, for a fixed \(b\),
\begin{equation}
\mathrm{AUC}_{\text{row}} \;=\; 
\frac{1}{n} \sum_{a=0}^{n-1} \frac{H[a,b] + H[a+1,b]}{2},
\end{equation}
and symmetrically, for a fixed \(a\),
\begin{equation}
\mathrm{AUC}_{\text{col}} \;=\; 
\frac{1}{m} \sum_{b=0}^{m-1} \frac{H[a,b] + H[a,b+1]}{2}.
\end{equation}
Both $\mathrm{AUC}_{\text{row}}$ and $\mathrm{AUC}_{\text{col}}$ are normalized to lie in \([0,1]\), 
providing a smooth global measure of invariance along each axis separately.


\medskip
\noindent \textbf{Baseline models.} To evaluate the embedding LLM base model $\phi$ (see Equation~\ref{eq:LLM_embedding_function}), we select a diverse suite of models, which collectively illustrate a larger spectrum of Transformers \cite{vaswani2017attention} architecture implementations available. This selection allows us to investigate whether the Platonic convergence of structural invariants is a universal property of Transformers or dependent on specific training paradigms:

\begin{itemize}
    \item \textbf{Commercial closed-source decoder-only models:} We include \textit{GPT-3.5} \cite{achiam2023gpt}, representing the industry standard for proprietary black-box embeddings in RAG pipelines.
    
    \item \textbf{Open-source decoder-only families:} We evaluate a spectrum of high-performance models including \texttt{Gemma~2} \cite{team2024gemma}, \texttt{Llama~3} \cite{grattafiori2024llama}, and the reasoning-optimized \texttt{DeepSeek-R1} \cite{liu2024deepseek}. These represent the state-of-the-art in autoregressive representation learning.
    
    \item \textbf{Specialized encoder-Only architectures:} We utilize the lightweight \texttt{nomic-embed-text} \cite{nussbaum2024nomic}, a model specifically optimized for long-context retrieval tasks.
    
    \item \textbf{TRL-refined baselines ($\Psi_{\theta}$):} We compare these general-purpose LLMs against a specialized structure-aware TRL encoder \cite{TCHUITCHEU2024110734}. This model uses a refined Transformer-based architecture \cite{vaswani2017attention} with three layers and 12 attention heads ($d=768$), serving as a structure-aware control group.
\end{itemize}




\begin{figure*}[t]
\begin{minipage}[b]{1.0\linewidth}
  \centering
  \centerline{\includegraphics[height=4cm, width=1\textwidth]{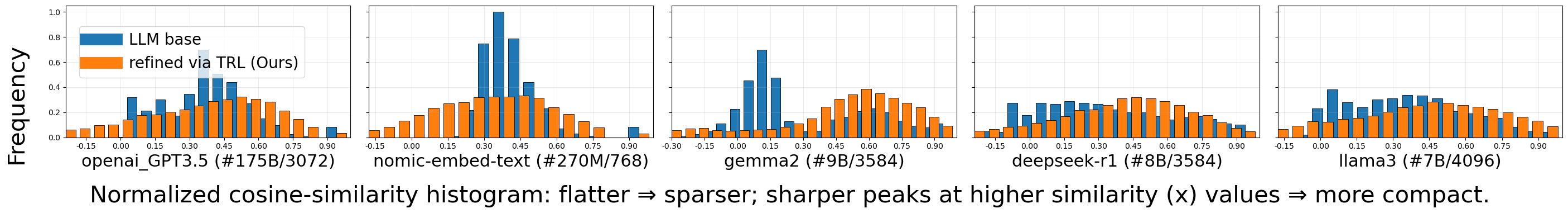}}
  \caption{Normalized histograms of pairwise cosine similarities for cell-cell embeddings: LLM-only (blue) vs.\ TRL-refined (orange) across five base models. LLM-only spaces exhibit narrow peaks at moderate similarity (\(\sim 0.30\!-\!0.50\)), whereas TRL refinement produces
    broader, flatter distributions with more mass up to \(\sim 0.75\). The TRL-refined model yields more consistent histogram shapes across models, indicating better cross-model compatibility.}\medskip
  \label{fg:cosinmilary_embeddingspace}
\end{minipage}
\begin{minipage}[b]{.32\linewidth}
  \centering
  \centerline{\includegraphics[width=1\textwidth]{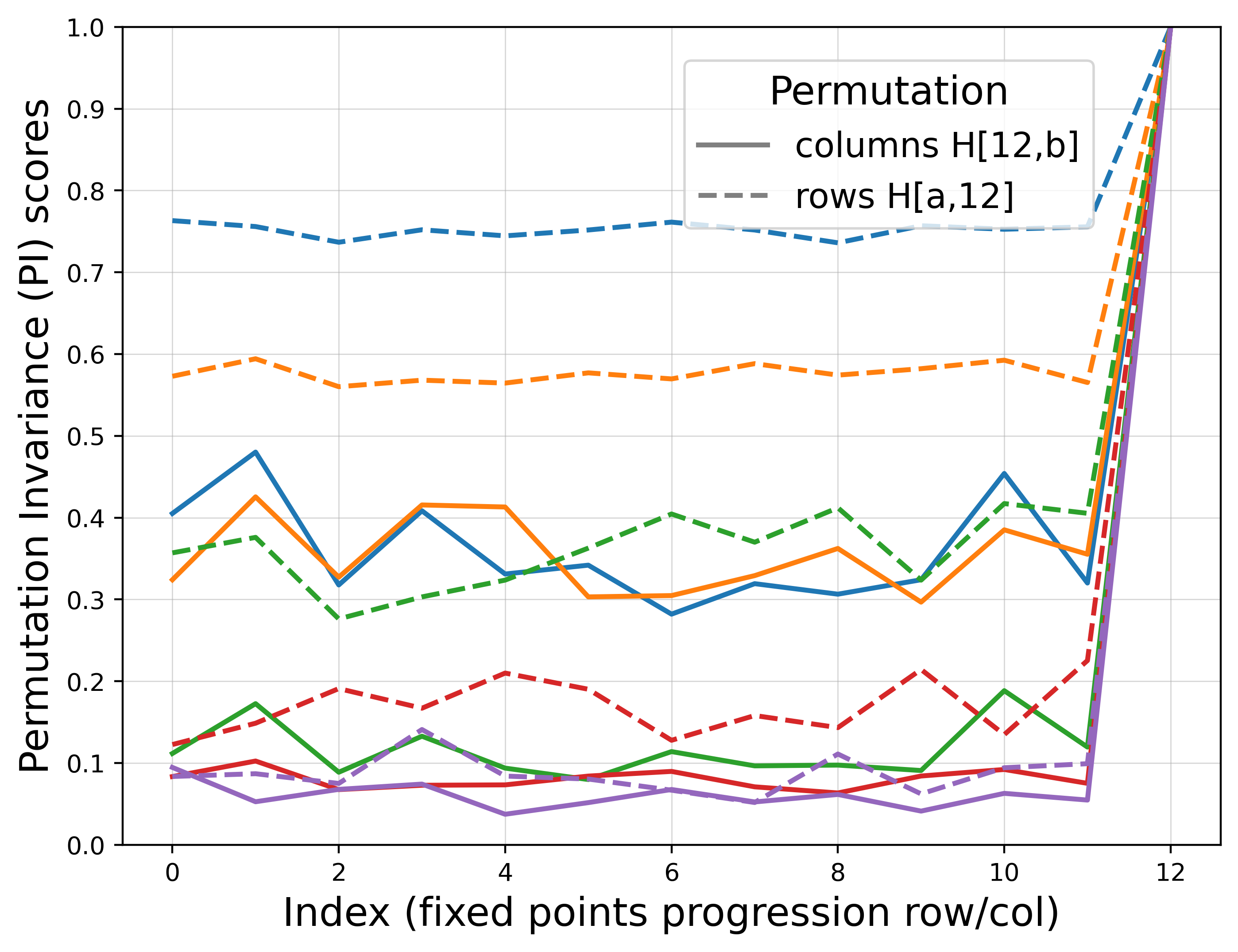}}
  \centerline{(b) LLM embeddings }\medskip
\end{minipage}
\hfill
\begin{minipage}[b]{0.32\linewidth}
  \centering
  \centerline{\includegraphics[width=1\textwidth]{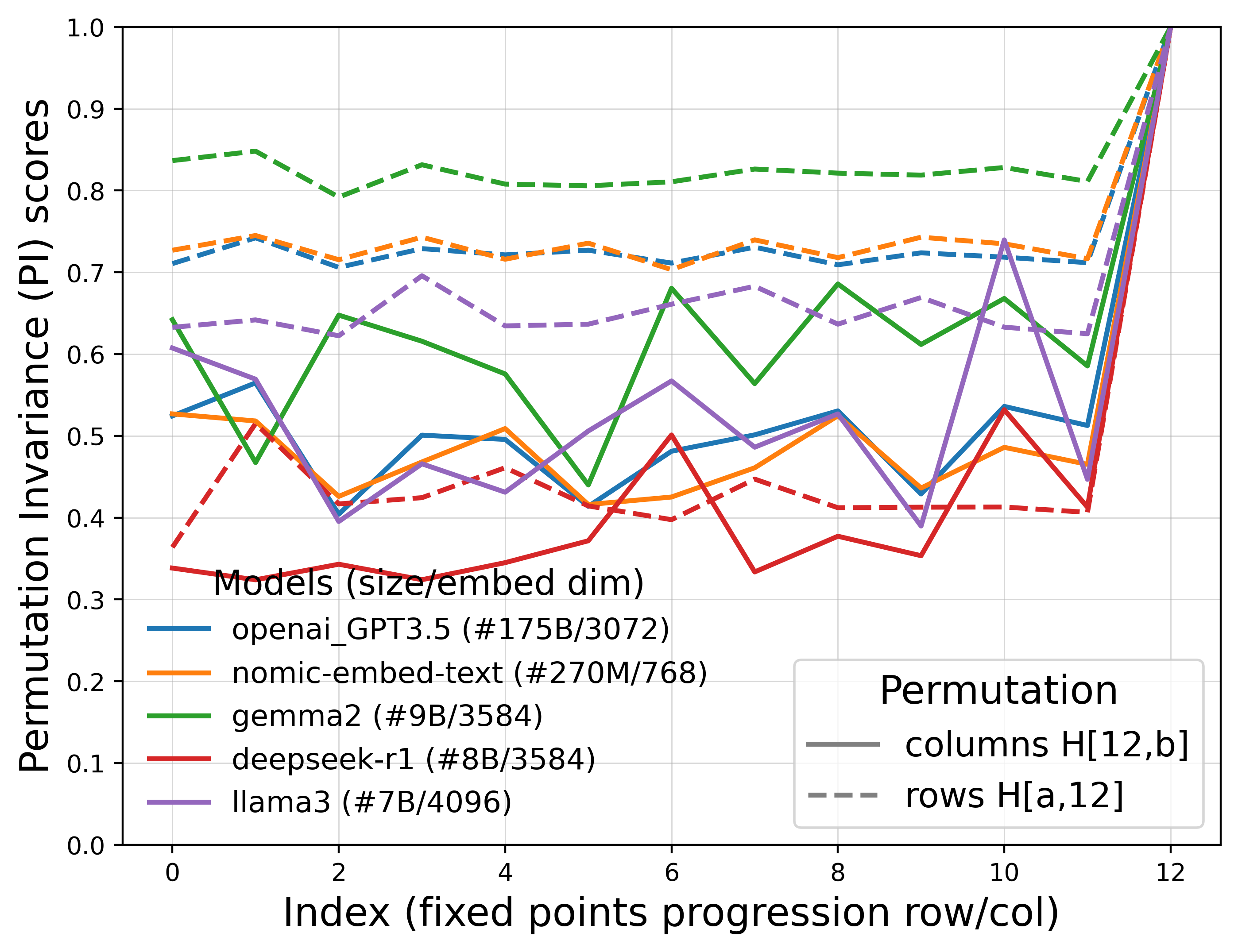}}
  \centerline{(c) TRL-refined embeddings (intra model)}\medskip
\end{minipage}
\hfill
\begin{minipage}[b]{0.32\linewidth}
  \centering
  \centerline{\includegraphics[width=1\textwidth]{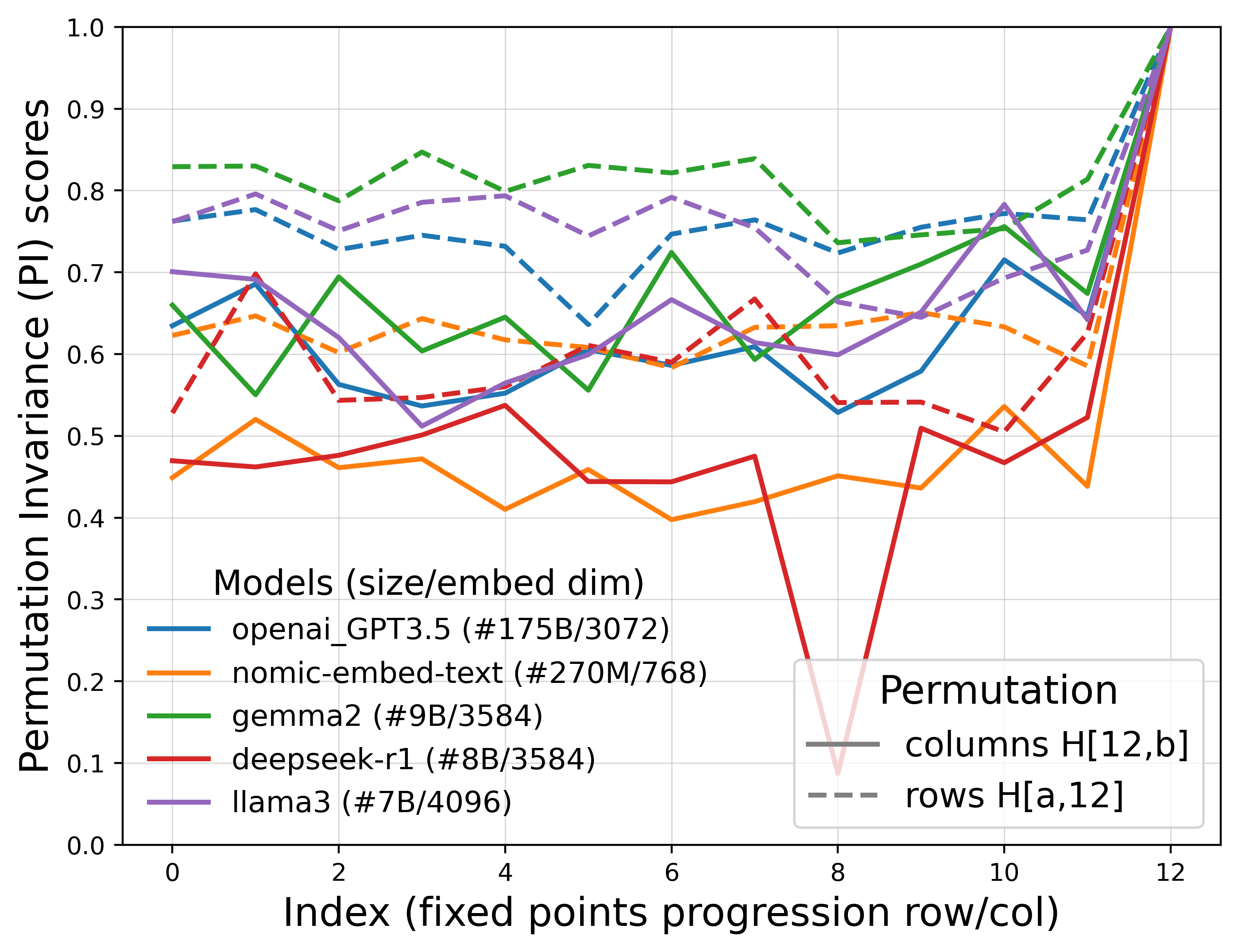}}
  \centerline{(c) TRL-refined embeddings (cross model)}\medskip
\end{minipage}
\caption{Analysis of the Platonic Representation Hypothesis (PRH) under progressive row and column shuffling. Each curve illustrates the similarity score \( H[a,b] \) as the number of fixed rows (\( a \)) or columns (\( b \)) increases, ranging from full derangement (\( T_0^0 \)) to complete identity (\( T_n^m \)). Solid lines represent column permutations (\( H[12,b] \)), while dashed lines represent row permutations (\( H[a,12] \)). The three panels compare: (left) baseline LLM embeddings, (center) TRL-refined embeddings (intra-model), and (right) TRL-refined embeddings (cross-model). The legend indicates the model size and embedding dimension. The TRL curves are flatter and more stable, demonstrating stronger PI, whereas the LLM-only curves exhibit a steeper decline, particularly under column permutations.}

\label{fig:LLMVDours_cos_row_column_effect}
\end{figure*}

\subsection{Evaluation on single-table}
\label{subsec:single}
Our first benchmark uses \(169\) permuted tables \(T_a^b\) derived from the baseline table shown in Figure~\ref{tab:rugby_stats} where \(n{=}12\) rows and \(m{=}12\) columns. For each model, we extract per–cell embeddings (\(\Phi\) for LLM baselines via Equation~\ref{eq:LLM_embedding_function}; \(\Psi_\theta\) for TRL), and:
(i) visualize cell embeddings of the original table (left column of Figure \ref{fig:LLM-based embedding}); (ii) the pairwise cosine similarity shown in the center of Figure \ref{fig:LLM-based embedding};
(iii) build the \emph{Platonic heatmap} \(H\) shown in the right column of Figure \ref{fig:LLM-based embedding}.
\medskip

\noindent \textbf{Analysing cell embeddings visualization.} \texttt{openai\_GPT3.5} and \texttt{nomic-embed-text} exhibit clear \emph{cell-type} clustering (headers vs.\ data cells), whereas other models do not, highlighting that different LLMs induce qualitatively different embedding geometries and cluster structure.

\medskip
\noindent  \textbf{Analysing embedding space geometric.} The within-table pairwise cosine-similarity patterns vary substantially across models, indicating that the relative neighborhood structure among cells is model-dependent. This suggests that the resulting embedding spaces are \emph{not} readily comparable (and not isometric in practice), which complicates cross-model alignment. To contrast LLM-only embeddings with structure-aware TRL embeddings, Figure~\ref{fg:cosinmilary_embeddingspace} reports normalized (to $[0,1]$) pairwise cosine-similarity matrices for both approaches. Compared with LLM-only baselines, TRL-refined embeddings exhibit a more coherent similarity structure (e.g., clearer block patterns and more consistent density), consistent with the hypothesis that injecting an equivariant table structure during refinement yields more stable representations.

\medskip
\noindent \textbf{Analysing the CKA heatmap.} The right panel of Figure~\ref{fg:cosinmilary_embeddingspace} reports the CKA heatmap \(H\) between the original table and its permuted variants. As expected, \(H[12,12]=1\), corresponding to the identity permutation \(T_{12}^{12}\). Along the rightmost column (\(a<12\), i.e., row-only permutations), \(H[a,12]\) remains high---particularly for \texttt{openai\_GPT3.5}, \texttt{nomic-embed-text}, and \texttt{gemma2}---indicating relatively low sensitivity to row shuffles. In contrast, any column permutation (\(b<12\)) substantially reduces \(H[12,b]\), suggesting that LLM-only embeddings are markedly more sensitive to column order than to row order. To compare LLM-only embeddings with TRL-refined embeddings, we flatten each model’s heatmap values and summarize their distributions using boxplots (Figure~\ref{fig:cka-boxplot_LLM-embed}). From this perspective, TRL-derived embeddings consistently achieve higher CKA scores, aligning more closely with the PRH. This improvement is consistent with the structural awareness introduced during refinement (e.g., explicit header--cell alignment).




\begin{table*}[!t]
\centering

\caption{Average performance (mean $\pm$ std.) over 3{,}791 permuted tables generated from 20 larger base tables with \(n \in [10,30]\) rows and \(m \in [10,15]\) columns. We evaluate each model under row- and column-permutation settings derived from the heatmap \(H[a,b]\), and report results for both base LLM embeddings and their TRL-refined counterparts, highlighting gains in permutation robustness.
}

\label{tab:report_large_eval}
\begin{adjustbox}{width=1\textwidth,center}
\resizebox{\linewidth}{!}{%
\begin{tabular}{@{}lcccccccc@{}}
\toprule
Models           & \multicolumn{3}{c|}{$\rho_{\text{mono}}$}                & \multicolumn{3}{c|}{$\mathrm{PI}_{\text{derange}}$}   & \multicolumn{2}{c}{AUC}  \\ \midrule
\multicolumn{1}{c}{} &
  \begin{tabular}[c]{@{}c@{}}rows\\ H{[}a,0{]}\end{tabular} &
  \begin{tabular}[c]{@{}c@{}}cols\\ H{[}0,b{]}\end{tabular} &
  \multicolumn{1}{c|}{\begin{tabular}[c]{@{}c@{}}all\\ H{[}a,b{]}\end{tabular}} &
  \begin{tabular}[c]{@{}c@{}}rows\\ H{[}a,0{]}\end{tabular} &
  \begin{tabular}[c]{@{}c@{}}cols\\ H{[}0,b{]}\end{tabular} &
  \begin{tabular}[c|]{@{}c@{}}all\\ H{[}a,b{]}\end{tabular} & 
  \begin{tabular}[c]{@{}c@{}}rows\\ H{[}a,0{]}\end{tabular} & \begin{tabular}[c]{@{}c@{}}Cols\\ H{[}0,b{]}\end{tabular} \\ \midrule
\multicolumn{9}{c}{LLM base Embedding}                                                                                             \\ \midrule
openai\_GPT3.5   & 0.23 ± 0.19 & 0.20  ± 0.23                & \multicolumn{1}{c|}{0.17 ± 0.06} & \textbf{0.77 ±0.13}                  & 0.42 ±0.17                 & \multicolumn{1}{c|}{0.38 ± 0.17} & 0.78 ± 0.12 & 0.40 ± 0.21 \\
nomic-embed-text & 0.29 ± 0.21 & 0.22 ± 0.26                  & \multicolumn{1}{c|}{0.20 ± 0.07} & 0.61  ± 0.16                & 0.39   ± 0.16               & \multicolumn{1}{c|}{0.35 ±0.16} & 0.63 ± 0.15 & 0.37 ± 0.19 \\
gemma2           & 0.20 ± 0.21 & 0.18 ±0.14                 & \multicolumn{1}{c|}{0.18 ± 0.1} & 0.56 ± 0.16                  & 0.22 ± 0.11                 & \multicolumn{1}{c|}{0.18 ± 0.09} & 0.60 ± 0.12 & 0.22 ± 0.14 \\
deepseek-r1      & 0.23 ± 0.29 & 0.32 ± 0.18                  & \multicolumn{1}{c|}{0.17 ± 0.07} & 0.34 ± 0.14                  & 0.10  ± 0.07                & \multicolumn{1}{c|}{0.11 ±0.1} &  0.38 ±0.13 & 0.15±0.14 \\
llama3           & 0.23 ± 0.25    & 0.16 ± 0.31                    & \multicolumn{1}{c|}{0.12 ± 0.14}    & 0.32 ±0.18                     & 0.19   ± 0.13                 & \multicolumn{1}{c|}{0.19 ± 0.1}  & 0.38 ± 0.17 & 0.22 ± 0.16 \\ \midrule
\multicolumn{9}{c}{ TRL-refined LLM embedding via TRL (intra-model)}                                                                \\ \midrule
openai\_GPT3.5   & 0.26 ± 0.21   & \multicolumn{1}{l}{0.31 ± 0.2} & \multicolumn{1}{l|}{0.2 ± 0.08}    & \multicolumn{1}{l}{0.73 ± 0.06} & \multicolumn{1}{l}{0.52 ± 0.05} & \multicolumn{1}{c|}{0.49 ± 0.08} & 0.75 ± 0.05 & 0.49 ± 0.19   \\
nomic-embed-text & 0.29 ± 0.21    & \multicolumn{1}{l}{\textbf{0.33 ± 0.19}} & \multicolumn{1}{l|}{0.18 ± 0.06}    & \multicolumn{1}{l}{0.74 ± 0.08} & \multicolumn{1}{l}{0.53 ± 0.07} & \multicolumn{1}{c|}{\textbf{0.56 ± 0.14}} & 0.77 ± 0.07 & 0.49 ± 0.19 \\
gemma2           & 0.27 ± 0.22   & \multicolumn{1}{l}{0.25 ± 0.19} & \multicolumn{1}{l|}{0.23 ± 0.09}    & \multicolumn{1}{l}{\textbf{0.77 ± 0.12}} & \multicolumn{1}{l}{\textbf{0.61 + 0.14}} & \multicolumn{1}{c|}{0.55 ± 0.13}  & \textbf{0.79 ± 0.1} & \textbf{0.56 ± 0.26}  \\
deepseek-r1      & \textbf{0.31 ± 0.27}    & \multicolumn{1}{l}{0.29 ± 0.19} & \multicolumn{1}{l|}{\textbf{0.25 ± 0.09}}    & \multicolumn{1}{l}{0.60 ± 0.12} & \multicolumn{1}{l}{0.49 ± 0.12} & \multicolumn{1}{c|}{0.44 ± 0.11}  & 0.63 ± 0.1 & 0.46 ± 0.23 \\
llama3           & 0.30 ± 0.23   & \multicolumn{1}{l}{0.31 ± 0.23} & \multicolumn{1}{l|}{0.23 ± 0.07}    & \multicolumn{1}{l}{0.69 ± 0.08} & \multicolumn{1}{l}{0.55 ± 0.1} & \multicolumn{1}{c|}{0.51 ± 0.09}  & 0.71 ± 0.08 & 0.51 ± 0.21  \\ \bottomrule
\end{tabular}
}
\end{adjustbox}
\end{table*}

\medskip 

\noindent\textbf{Ablation: row vs.\ column permutations.} We isolate row- and column-permutation effects by analyzing the two heatmap slices in Figure~\ref{fig:LLMVDours_cos_row_column_effect}b--c): \(H[a,12]\) (column order fixed; rows permuted) and \(H[12,b]\) (row order fixed; columns permuted). We summarize each slice with two scalars:
(i) an \emph{axis-derangement score},
\(\mathrm{PI}_{\mathrm{derange}}=H[0,12]\) for rows (or \(H[12,0]\) for columns), capturing invariance under full derangement along one axis; and
(ii) a monotonicity coefficient,
\(\rho_{\mathrm{mono}}=\operatorname{Spearman}(\{H[a,12]\},\{a\})\) (or \(\operatorname{Spearman}(\{H[12,b]\},\{b\})\)),
which quantifies how consistently similarity recovers as fixed rows/columns are restored toward \(H[12,12]=1\).

Across LLM-only embeddings we observe \(\mathrm{PI}_{\mathrm{derange}}\in[0.08,0.76]\) and \(\rho_{\mathrm{mono}}\in[-0.06,0.66]\), whereas TRL-refined embeddings attain higher values, \([0.34,0.94]\) and \([-0.02,0.79]\), respectively.
For example, the strongest LLM baseline (\texttt{openai\_GPT3.5}) achieves \(0.76/0.03\) under row permutations, while TRL reaches \(0.94/0.02\), reflecting a higher overall similarity profile and greater permutation robustness, consistent with closer adherence to the PRH.

Interestingly, among LLM baselines, the 270M-parameter \texttt{nomic-embed-text} consistently rivals a 175B-parameter model, suggesting that \emph{embedding dimensionality} and \emph{training objectives} may influence PI more than sheer parameter count.
Finally, in the embedding space (Figure~\ref{fig:LLMVDours_cos_row_column_effect}a), TRL embeddings exhibit more similar global geometry across models, whereas the best-performing LLM baseline forms a noticeably denser configuration.

\subsection{Evaluation on multiple tables}
\label{subsec:multi}

We further conduct experiments on a benchmark of 3{,}791 tables generated from 20 base tables with sizes
\(n\in[10,30]\) (rows) and \(m\in[10,15]\) (columns). We report the mean
\(\mathrm{PI}_{\text{derange}}\) and \(\rho_{\text{mono}}\) along with the standard deviation in
Table~\ref{tab:report_large_eval}. We provide a detailed analysis of Table \ref{tab:report_large_eval} for each metric. 

\medskip
\noindent\textbf{Monotonicity coefficient $\rho_{\text{mono}}$:}  
  Measures whether similarity $H[a,b]$ increases monotonically as more rows ($a$) or columns ($b$) are restored from derangement back to identity. For LLM baselines, $\rho_{\text{mono}}$ is slightly lower (e.g., $0.17$ overall for OpenAI \texttt{GPT-3.5}, $0.12$ for \texttt{Llama3}), 
  indicating that embeddings fail to recover consistently with the structure.  
  In contrast, TRL-refined embeddings systematically improve (up to $0.31$ for \texttt{DeepSeek-r1}), showing stronger alignment with the PRH.
  
\medskip
 \noindent \textbf{Permutation Invariance $\mathrm{PI}_{\text{derange}}$:}  
  quantifies how much embeddings degrade under full derangement relative to identity.  Larger values indicate more invariance to shuffling.  
  LLMs show sharp drops, especially under column permutations 
  (e.g., \texttt{GPT-3.5}: $0.42$ for columns vs.\ $0.77$ for rows), confirming that column order is more affected by the permutation than row order.  
  TRL models achieve consistently higher $\mathrm{PI}_{\text{derange}}$ scores  (e.g., \texttt{Gemma2} improves from $0.22$ to $0.61$ on columns),  
  reflecting greater robustness to permutations.  
  
\medskip
 \noindent \textbf{Normalized AUC:}  
  captures the smoothness of recovery across the full permutation trajectory from $T_0^0$ (full shuffle) to $T_n^m$ (identity).  
  For LLMs, AUC remains low under column permutations 
  (e.g., \texttt{Llama3}: $0.22$), reflecting poor global performance.  
  TRL refinement substantially increases AUC (up to $0.56$ for \texttt{Gemma2} and $0.49$ for \texttt{GPT-3.5}), indicating that structural information encoded via TRL supports smoother, more reliable recovery.  

\medskip
\noindent\textbf{Key Findings.}  
Across all metrics, LLM-derived embeddings are more sensitive to column than row permutations, revealing a fundamental weakness in PRH.  By contrast, TRL consistently improves $\rho_{\text{mono}}$, $\mathrm{PI}_{\text{derange}}$, and AUC, demonstrating more stable invariance curves and stronger empirical support for PRH.  Notably, even smaller models (e.g., \texttt{nomic-embed-text}, $270$M parameters) benefit from TRL and can rival or exceed the permutation robustness of very large LLMs (e.g., \texttt{GPT-3.5}, $175$B parameters).  This highlights that robustness is driven less by raw model size and more by embedding structure and training objectives. 

\subsection{Limitations of the methodology}

Our analysis focuses on controlled row and column permutations within fully structured tables, which we define using a symmetric group action on tables \( h \) (see Equation~\ref{goup_action}). However, in real documents, tables often feature merged cells and hierarchical or nested headers, which contradict the assumptions required for a clean row/column-group action. This discrepancy renders the definition of the group action \( h \) ill-posed, complicating our evaluation of the proposed PRH. Therefore, extending PRH to accommodate such irregularities is an important direction for future research.

Finally, if PI is truly necessary for claiming semantic faithfulness—or for improving table retrieval—this must be established end-to-end rather than inferred from intrinsic diagnostics alone. Concretely, it requires a benchmark that \emph{pairs each table with controlled row/column-permuted variants} and evaluates whether retrieval and downstream reasoning remain consistent across these semantically equivalent inputs. Since such an end-to-end dataset and protocol are not yet available, constructing them is left for future work. We position this as part of the prospective vision this paper aims to push: making invariance-based evaluation a standard axis for next-generation table retrieval and table-RAG systems.

\subsection{Conclusion from the serialization-bias diagnostic}
\label{sec:conclusion}

We instantiate the \emph{Platonic Representation Hypothesis} (PRH) for tables and propose two permutation-based invariance metrics, \(\mathrm{PI}_{\text{derange}}\) and \(\rho_{\text{mono}}\), derived from CKA similarity. Applying these diagnostics shows that permutation sensitivity is not uniform across embedders. Several open-source models (e.g., \texttt{DeepSeek-r1}, \texttt{gemma2}, \texttt{llama3}) exhibit different sensitivity, while \texttt{openai\_GPT3.5} is comparatively more stable under \emph{row} permutations. Across all evaluated models, however, we observe clear semantic drift under \emph{column} permutations: different column orders of the same table can map far apart in embedding space, causing semantically equivalent tables to be treated as distinct retrieval candidates and weakening empirical support for PRH in current LLM-based embeddings.

In contrast, structure-aware TRL representations, anchored in header--cell correspondences, consistently yield higher invariance and reduced drift. Beyond reporting these findings, our metrics provide practical diagnostics for table-centric RAG: they can be used to compare embedders and to estimate how robust table retrieval will be under layout changes.

\medskip
\noindent \textbf{Recommendation.} For table-centric RAG, commercial embedders such as \texttt{openai\_GPT3.5} can be strong choices when cost permits, while lightweight open-source embedders such as \texttt{nomic-embed-text} provide a competitive alternative. When robustness to layout changes is a priority, TRL-refined embeddings are preferred; however, table-specific TRL methods remain less mature than the broader LLM embedding ecosystem.

\section{Prospective Vision: Toward Permutation-Invariant Retrieval}
\label{sec:prospective}

The diagnostic results above suggest a concrete bottleneck for table-centric IR: under common serialization choices, embeddings can drift substantially under benign permutations, especially column re-ordering. This motivates a prospective shift in emphasis from improving table-RAG via prompting and generation alone to improving it at the \emph{representation layer}, by explicitly targeting structural invariants. In this view, ``Platonic representations'' are not a claim that a universal embedding already exists, but an agenda: define invariance requirements that table representations should satisfy and evaluate retrieval systems against them. We outline two research directions that follow from this agenda.

First, we propose \textbf{Online Table Question Answering (OTQA)} as a potential downstream use-case of invariant-aware embeddings. If a representation is stable under table permutations, then a subset of queries---such as point lookups and lightweight relational checks---may be answerable through fast retrieval and simple geometric operations in embedding space (e.g., nearest-neighbor selection under constraints), reducing reliance on repeated prompting and generation. This direction is conditional on stronger invariance than current generic embeddings provide, and therefore serves as a clear target for future table representation learning.

Second, as industrial IR increasingly requires \textbf{complex relational operations}---including multi-step filtering, aggregation, and cross-table linking---the need for a standardized, structure-compatible representation becomes more pressing. Our findings indicate that column-order sensitivity can undermine retrieval reliability even before reasoning begins, suggesting that future systems will benefit from embeddings that are stable under benign table transformations. A practical implication is the need for end-to-end benchmarks that include controlled permuted variants of the same table and evaluate whether retrieval and downstream reasoning remain consistent. Establishing such protocols would help make PI a standard evaluation axis for table retrieval and table-RAG pipelines.

Finally, invariant-aware representations connect directly to three operations that are currently fragile in linearized table-RAG pipelines. \textit{Multi-step aggregation} should not depend on row order and therefore benefits from row-stable representations. \textit{Relational filtering} requires capturing relationships (e.g., header--cell semantics) beyond local lexical overlap. \textit{Cross-table referencing} requires identifying compatible columns or entities across heterogeneous tables despite schema differences. These directions are prospective, but they are grounded in the failure modes revealed by our diagnostics and provide concrete targets for future work on permutation-invariant retrieval.

\section{Conclusion}
\label{sec:conclusion}

In this prospective paper, we have explored the critical intersection of Table Representation Learning (TRL) and the Retrieval-Augmented Generation (RAG) landscape through the lens of the \textit{Platonic Representation}. We empirically demonstrated that the ubiquitous reliance on table linearization in current Information Retrieval (IR) pipelines can introduce a destructive bias, dissolving the essential structural invariants—specifically row-column PI—that define tabular data. This structural dissolution persists even in models specifically fine-tuned for tabular tasks, highlighting a fundamental need to rethink how tables are learned.

By positioning Platonic representations as a necessary paradigm shift, we outlined a dual-pathway vision for the future of IR. We introduced the concept of \textbf{Online Table Question Answering (OTQA)}, demonstrating how invariant embeddings can facilitate direct, "prompt-less" reasoning within the latent space, thereby drastically reducing the computational overhead of RAG. Furthermore, we detailed how these robust representations provide the "structural fuel" for complex industrial operations, including multi-step aggregation and semantic cross-table referencing. 

Finally, we posit that the quest for the "Platonic representation" of structured data is not merely a theoretical exercise but a practical prerequisite for the next generation of intelligent systems. As the IR field moves toward a RAG-centric future, adopting a standardized, invariant representation for tables will be the cornerstone that transforms generative models from surface-level retrievers into deep analytical engines.

\section{Acknowledgments}
This research was supported by funding from the Flemish Government under the “Onderzoeksprogramma Artificiële Intelligentie (AI) Vlaanderen” program.

\newpage
\balance
\bibliographystyle{ACM-Reference-Format}
\bibliography{sample-base}
\end{document}